\documentclass[10pt,twocolumn,letterpaper]{article}

\usepackage{cvpr}
\usepackage{times}
\usepackage{epsfig}
\usepackage{listings}
\usepackage[dvipsnames]{xcolor}

\definecolor{codegreen}{rgb}{0,0.6,0}
\definecolor{codegray}{rgb}{0.5,0.5,0.5}
\definecolor{codepurple}{rgb}{0.58,0,0.82}
\definecolor{backcolour}{rgb}{0.95,0.95,0.92}

\lstdefinestyle{mystyle}{
    backgroundcolor=\color{backcolour},   
    commentstyle=\color{codegreen},
    keywordstyle=\color{magenta},
    numberstyle=\tiny\color{codegray},
    stringstyle=\color{codepurple},
    basicstyle=\ttfamily\footnotesize,
    breakatwhitespace=false,         
    breaklines=true,                 
    captionpos=b,                    
    keepspaces=true,                 
    numbers=left,                    
    numbersep=5pt,                  
    showspaces=false,                
    showstringspaces=false,
    showtabs=false,                  
    tabsize=2
}

\usepackage{graphicx}
\usepackage{amsmath}
\usepackage{amssymb}
\usepackage{enumitem}
\usepackage{bbm}
\usepackage{multirow}
\usepackage{caption} 
\usepackage{array}
\usepackage{arydshln}
\usepackage[square,numbers]{natbib}
\usepackage{subcaption}

\makeatletter
\newcommand{\printfnsymbol}[1]{%
  \textsuperscript{\@fnsymbol{#1}}%
}
\usepackage{amsmath,amsfonts,bm}

\def\eqref#1{equation~\ref{#1}}
\def\1{\bm{1}}

\def\rvalpha{{\mathbf{\alpha}}}
\def\rvpi{{\mathbf{\pi}}}
\def\rva{{\mathbf{a}}}
\def\rvb{{\mathbf{b}}}
\def\rvc{{\mathbf{c}}}

\def\rvf{{\mathbf{f}}}
\def\rvg{{\mathbf{g}}}
\def\rvh{{\mathbf{h}}}
\def\rvu{{\mathbf{i}}}

\def\rvs{{\mathbf{s}}}

\def\rvu{{\mathbf{u}}}
\def\rvv{{\mathbf{v}}}

\def\rvx{{\mathbf{x}}}
\def\rvy{{\mathbf{y}}}
\def\rvz{{\mathbf{z}}}

\def\rmF{{\mathbf{F}}}

\def\rmH{{\mathbf{H}}}

\def\rmP{{\mathbf{P}}}

\def\rmR{{\mathbf{R}}}

\def\rmV{{\mathbf{V}}}
\def\rmW{{\mathbf{W}}}
\def\rmX{{\mathbf{X}}}

\DeclareMathAlphabet{\mathsfit}{\encodingdefault}{\sfdefault}{m}{sl}
\SetMathAlphabet{\mathsfit}{bold}{\encodingdefault}{\sfdefault}{bx}{n}

\newcommand{\R}{\mathbb{R}}

\newcommand{\softmax}{\mathrm{softmax}}

\DeclareMathOperator*{\argmax}{arg\,max}

\newcommand{\norm}[1]{\left\lVert#1\right\rVert}

\DeclareMathOperator{\simop}{sim}
\def\Z{{\mathcal{Z}}}
\def\U{{\mathcal{U}}}

\usepackage{arydshln}

\usepackage[pagebackref=true,breaklinks=true,letterpaper=true,colorlinks,bookmarks=false]{hyperref}
\usepackage[capitalise]{cleveref}                 % Best reference package
\crefname{equation}{}{}                           % Removing eq.
\Crefname{equation}{}{}                           % Removing Eq.

\cvprfinalcopy % *** Uncomment this line for the final submission

\def\cvprPaperID{2866} % *** Enter the CVPR Paper ID here

\begin{document}

%%%%%%%%% TITLE
\title{Sequence-to-Sequence Contrastive Learning for Text Recognition}

% \author{
%     Aviad Aberdam\thanks{Authors contribute equally and are listed in alphabetical order.} \printfnsymbol{2}\printfnsymbol{3}, Ron Litman\printfnsymbol{1}\printfnsymbol{2}, Shahar Tsiper\printfnsymbol{2}, Oron Anschel\printfnsymbol{2}, Ron Slossberg\printfnsymbol{2}\printfnsymbol{3},\\
%     Shai Mazor\printfnsymbol{2}, R. Manmatha\printfnsymbol{2}, and Pietro Perona\printfnsymbol{2} \\
%     \printfnsymbol{2}Amazon Web Services, \printfnsymbol{3} Technion -- Israel Institute of Technology\\
%     {\tt\small \{aaberdam, ronslos\}@cs.technion.ac.il}\\
%     {\tt\small \{litmanr, tsiper, oronans, smazor, manmatha, peronapp\}@amazon.com}
% }

\author{Aviad Aberdam\thanks{Authors contribute equally and are listed in alphabetical order.}\\
Technion\\
{\tt\footnotesize aaberdam@cs.technion.ac.il}
\and
Ron Litman\printfnsymbol{1}\\
AWS\\
{\tt\footnotesize litmanr@amazon.com}
\and
Shahar Tsiper\\
AWS\\
{\tt\footnotesize tsiper@amazon.com}
\and
Oron Anschel\\
AWS\\
{\tt\footnotesize oronans@amazon.com}
\and
Ron Slossberg\\
Technion\\
{\tt\footnotesize ronslos@cs.technion.ac.il}
\and
Shai Mazor\\
AWS\\
{\tt\footnotesize smazor@amazon.com}
\and
R. Manmatha\\
AWS\\
{\tt\footnotesize manmatha@amazon.com}
\and
Pietro Perona\\
Caltech and AWS\\
{\tt\footnotesize peronapp@amazon.com}
}

\maketitle
\newcommand{\AlgoName}{SeqCLR}

%%%%%%%%% ABSTRACT
\begin{abstract}
    We propose a framework for sequence-to-sequence contrastive learning (SeqCLR) of visual representations, which we apply to text recognition. To account for the sequence-to-sequence structure, each feature map is divided into different instances over which the contrastive loss is computed. This operation enables us to contrast in a sub-word level, where from each image we extract several positive pairs and multiple negative examples. To yield effective visual representations for text recognition, we further suggest novel augmentation heuristics, different encoder architectures and custom projection heads. Experiments on handwritten text and on scene text show that when a text decoder is trained on the learned representations, our method outperforms non-sequential contrastive methods. In addition, when the amount of supervision is reduced, SeqCLR significantly improves performance compared with supervised training, and when fine-tuned with 100\% of the labels, our method achieves state-of-the-art results on standard handwritten text recognition benchmarks.
\end{abstract}

%%%%%%%%% BODY TEXT
%-------------------------------------------------------------------------
\begin{figure}[t]
  \centering
  \includegraphics[width=0.9\columnwidth]{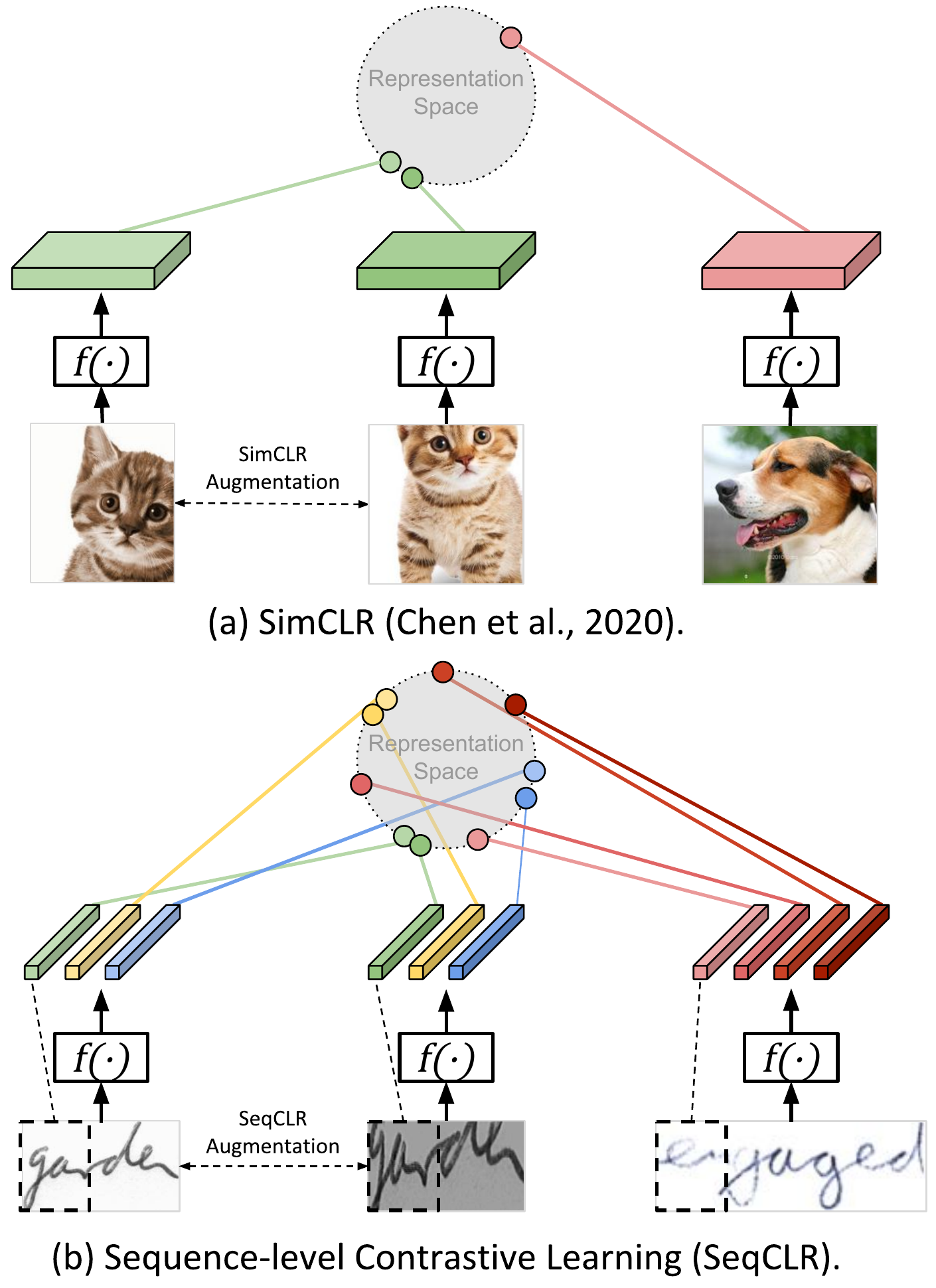}
  \caption{{\bf Sequence contrastive learning.} (a) Current contrastive methods compare representations computed from whole images. (b) We propose a sequence-to-sequence approach, by viewing the feature map as a sequence of separate representations. This is useful in text recognition, where words are composed of sequences of characters.}
  \label{fig:teaser_fig_mapping}
  \vspace{-0.3cm}
\end{figure}

\section{Introduction}
\label{sec:introduction}

Contrastive learning techniques for self-supervised representation learning have recently demonstrated significant improvements on several semi-supervised computer vision applications, including image classification, object detection, and segmentation~\cite{chen2020simple, he2020momentum, falcon2020framework, caron2020unsupervised, Han20, rao2020augmented,chen2020big,tschannen2020self, qian2020spatiotemporal}.
As illustrated in \cref{fig:teaser_fig_mapping}(a) contrastive learning is performed by maximizing agreement between representations of differently augmented views of the same image and distinguishing them from representations of other dataset images.

Despite obvious advantages, unsupervised and semi-supervised schemes have hardly been explored for text recognition (\cite{gupta2018learning, zhang2019sequence, kang2020unsupervised}). For example, currently, most handwritten text recognition approaches still rely on fully supervised learning, requiring large amounts of annotated data.
The reason for this is simple: current contrastive schemes for visual representation learning are tailored towards tasks such as object recognition or classification, where images are atomic input elements. For example, in image classification, positive examples are created by augmenting each image while all other images in the dataset are assumed to be negative (\cref{fig:teaser_fig_mapping}(a)). On the other hand, for sequential prediction as used in text recognition, a word is viewed as a sequence of characters, and thus the image of a word is best modeled as a sequence of adjacent image slices (frames), each one of which may represent a different class as depicted in \cref{fig:teaser_fig_mapping}(b). Thus, the standard `whole image' contrastive learning approach is inadequate for this task.

\begin{figure}[t]
\normalsize
  \centering
  \includegraphics[width=0.85\linewidth]{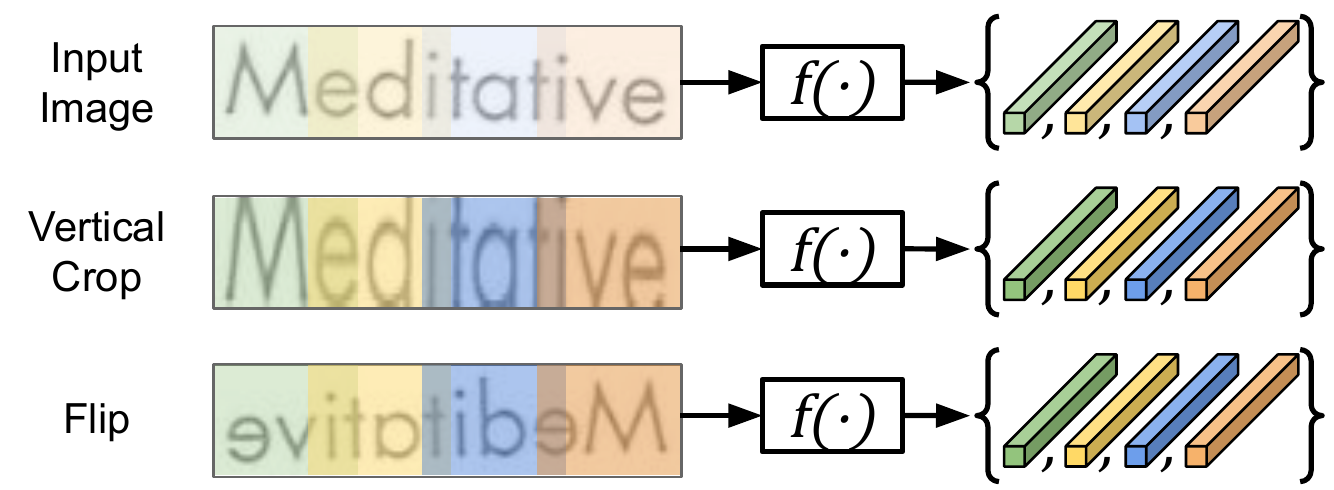}
  \caption{{\bf Sequence preserving augmentations.} We propose an augmentation procedure which meets the sequential structure of the feature map. For example, as opposed to vertical cropping, horizontal flipping results in a sequence-level misalignment which leads to poor positive pairing.}
  \label{fig:aug_receptive_field}
  \vspace{-0.3cm}
\end{figure}

We propose an approach that extends existing contrastive learning methods to sequential prediction tasks such as text recognition. The key idea is to apply contrastive learning to the individual elements of the sequence, while maintaining information about their order. To do so, we introduce an instance-mapping function that yields an instance from every few consecutive frames in a sequence feature map. The instance is the atomic element that will be used in contrastive learning. A given image, depending on its width, may produce an arbitrary number of instances. This enlarges the number of negative examples in every batch without requiring a memory bank~\cite{he2020momentum} or architecture modifications~\cite{bachman2019learning}.
Individual instances are part of a sequence, thus we design an augmentation procedure that ensures a sequence-level alignment, which is crucial for yielding effective representations (\cref{fig:aug_receptive_field}).

We validate our method experimentally, comparing its performance with non-sequential contrastive approaches on several handwritten and scene text datasets.
To evaluate the quality of the learned visual representation, we lay out a decoder evaluation protocol that extends the widely-used linear evaluation criteria~\cite{zhang2016colorful, kolesnikov2019revisiting} for encoder-decoder based networks. Utilizing this evaluation, we demonstrate significant improvements over current contrastive learning approaches.
Furthermore, we find that our method outperforms supervised training methods with limited amounts of labeled training data, and it achieves state-of-the-art results on standard handwritten datasets, reducing the word error rate by 9.5\% on IAM and by 20.8\% on RIMES.

To summarize, the key contributions of our work are:
\begin{itemize}
    \item A contrastive learning approach for visual sequence-to-sequence recognition.
    \item Viewing each feature map as a sequence of individual instances, leading to contrastive learning in a sub-word level, such that each image yields several positive pairs and multiple negative examples.
    \item Defining sequence preserving augmentation procedures, and custom projection heads.
    \item Extensive experimental validation showing state-of-the-art performance on handwritten text. 
\end{itemize}

%-------------------------------------------------------------------------

\section{Related Work}
\label{sec:related_work}

\paragraph{Visual representation learning}
Unsupervised representation learning has recently achieved success, not only in natural language processing~\cite{pennington2014glove, mikolov2013distributed,devlin2018bert,radford2018improving,radford2019language}
and speech recognition~\cite{baevski2020wav2vec, jiang2019improving, wang2020unsupervised},
but also in computer vision.
The first methods suggested learning visual representations by training the network on an artificially designed pretext task, such as denoising auto-encoders~\cite{vincent2008extracting}, patch ordering~\cite{doersch2015unsupervised}, colorizing an image~\cite{zhang2016colorful}, and others~\cite{noroozi2016unsupervised,gidaris2018unsupervised}.

In this paper, we focus on the contrastive learning approach, which has recently shown promising results on several tasks~\cite{hjelm2018learning, henaff2019data,bachman2019learning,chen2020simple,falcon2020framework, he2020momentum, chen2020big, caron2020unsupervised, rao2020augmented}.
In this method, we maximize agreement between representations of differently augmented views of the same data and contrast between representations coming from different images~\cite{bachman2019learning}.
This process may be viewed as a classification task where each image is assumed to be its own class.

Several papers explored this approach, introducing several advances over the base contrastive scheme.
The authors in~\cite{chen2020simple} proposed an augmentation pipeline and an additional projection head which maps the representations into space where the contrastive loss is applied.
In~\cite{he2020momentum} a momentum-based contrastive scheme was suggested, and~\cite{chen2020big} included a teacher-student distillation phase.
Additional papers~\cite{Han20, rao2020augmented, tschannen2020self, qian2020spatiotemporal} introduced contrastive learning schemes for action classification of sequential inputs and non-sequential outputs.
Motivated by these papers, we expand the contrastive learning framework to visual sequence-to-sequence predictions as in text recognition.

\begin{figure*}[t!]
\normalsize
  \centering
  \includegraphics[width=0.95\textwidth]{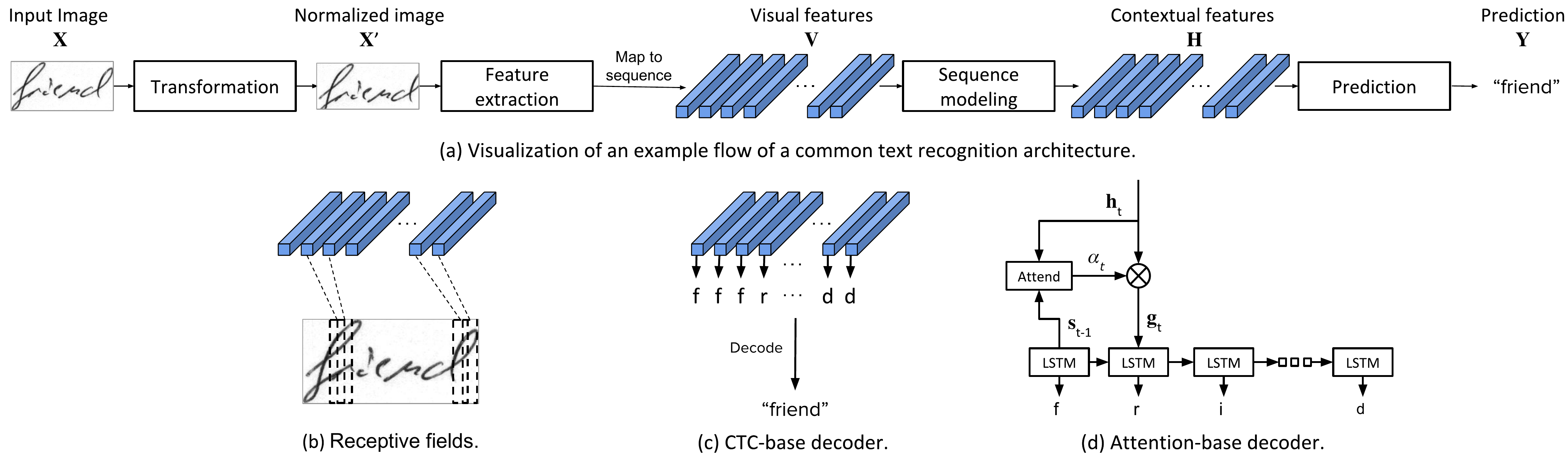}
  \caption{\textbf{Typical text recognition architecture}. Visualization of an example flow of a text recognition architecture (a), as described in Section \ref{sec:text_recog_background}. After the input text image is normalized, we extract a sequence of visual features from it, where each frame is associated with a different receptive field of the input image (b). Then, these visual representations go through a sequence modeling (LSTM scheme), and finally are decoded using a CTC (c) or an attention (d) decoder.}
  \label{fig:main_fig_text_recog}
  \vspace{-0.3cm}
\end{figure*}

\paragraph{Un- and semi-supervised learning for text recognition}
Despite clear advantages, currently, most text recognition methods do not utilize unlabeled real-world text images.
Specifically, handwritten recognition usually relies on fully-supervised training~\cite{yousef2020origaminet, sueiras2018offline}, while scene text models are trained mostly on synthetic data~\cite{baek2019wrong, litman2020scatter}.
That said,~\cite{zhang2019sequence} and~\cite{kang2020unsupervised} have recently suggested domain adaptation techniques to utilize an unlabeled dataset along with labeled data.
Using adversarial training, these methods align the feature map distributions of both datasets.

For images of printed text,~\cite{gupta2018learning} recently proposed a completely unsupervised scheme in which a discriminator enforces the predictions to align with a distribution of a given text corpus.
Nevertheless, this method requires restricting the recognizer architecture to use only local predictions.

To the best of our knowledge, this work is the first to propose self-supervised representation learning for text recognition. Our method further leads to state-of-the-art results on handwritten text.

%-------------------------------------------------------------------------
\section{Text Recognition Background}
\label{sec:text_recog_background}

Several architectures have been proposed over the years for recognition of scene text~\cite{Long2018survey, chen2020text, sengupta2020journey} and handwritten text~\cite{sonkusare2016survey, memon2020handwritten}.
Throughout this work, we focus on a general text recognition framework, which was proposed by~\cite{baek2019wrong}.
This framework describes the building blocks of many text recognizers, including~\cite{Bai2015crnn,Zisserman2015large,shi2016end,shi2016robust, liu2016star, Hu2017grccn,bai2017accurate,cheng2017focusing, fedor2018rosetta, zhang2019sequence, litman2020scatter, fogel2020scrabblegan, yousef2020origaminet}.
As shown in \cref{fig:main_fig_text_recog}, this architecture consists of the following four stages (see more details in \cref{app:text_recog_scheme}):
\begin{enumerate}
    \item \textbf{Transformation:} A normalization of the input text image using a Thin Plate Spline (TPS) transformation~\cite{shi2016robust, liu2016star}, which is a variant of the spatial transformer network~\cite{jaderberg2015spatial}. This stage is optional yet important for images of text in diverse shapes.
    \item \textbf{Feature extraction:} A convolutional neural network (CNN) that extracts features from the normalized image, followed by a map-to-sequence operation that reshapes the features into a sequence of frames, denoted by $\rmV = [\rvv_1, \rvv_2, \ldots, \rvv_T]$. As illustrated in \cref{fig:main_fig_text_recog}(b), the resulting frames correspond to different receptive fields in the image. Note that the sequence length depends on the width of the input image.
    \item \textbf{Sequence modeling:} An optional Bidirectional LSTM (BiLSTM) scheme which aims to capture the contextual information within the visual feature sequence. This network yields the contextual features $\rmH = [\rvh_1, \rvh_2, \ldots, \rvh_T]$, which in turn, are concatenated to the feature map $\rmV$, as suggested in~\cite{litman2020scatter}.
    \item \textbf{Prediction:} A text decoder using (i) a connectionist temporal classification (CTC) decoder~\cite{graves2006connectionist} that decodes separately each frame, and then deletes repeated characters and blanks (\cref{fig:main_fig_text_recog}(c)); or (ii) an attention decoder~\cite{cheng2017focusing, shi2016robust}, which linearly combines the frames to feed them into a one-layer LSTM (\cref{fig:main_fig_text_recog}(d)).
\end{enumerate}

\begin{figure*}[t]
\normalsize
  \centering
  \includegraphics[width=\textwidth]{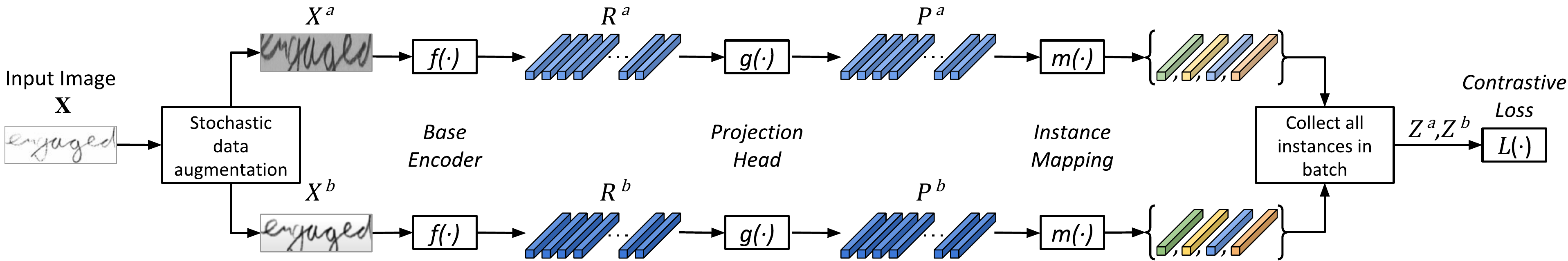}
  \caption{\textbf{\AlgoName{} block diagram.} Each image in a batch is augmented twice, and then fed separately into a base encoder and projection head, to create pairs of representation maps. Next, to account for the sequential structure of these representations, we apply an instance-mapping function (see also \cref{fig:insatnce_mapping_func}) that transforms them into several instances and thus allows us to apply contrastive learning at a sub-word level.
    }
  \label{fig:main_fig_framework}
  \vspace{-0.5cm}
\end{figure*}

\begin{figure}[t!]
\normalsize
  \centering
  \includegraphics[width=0.9\columnwidth]{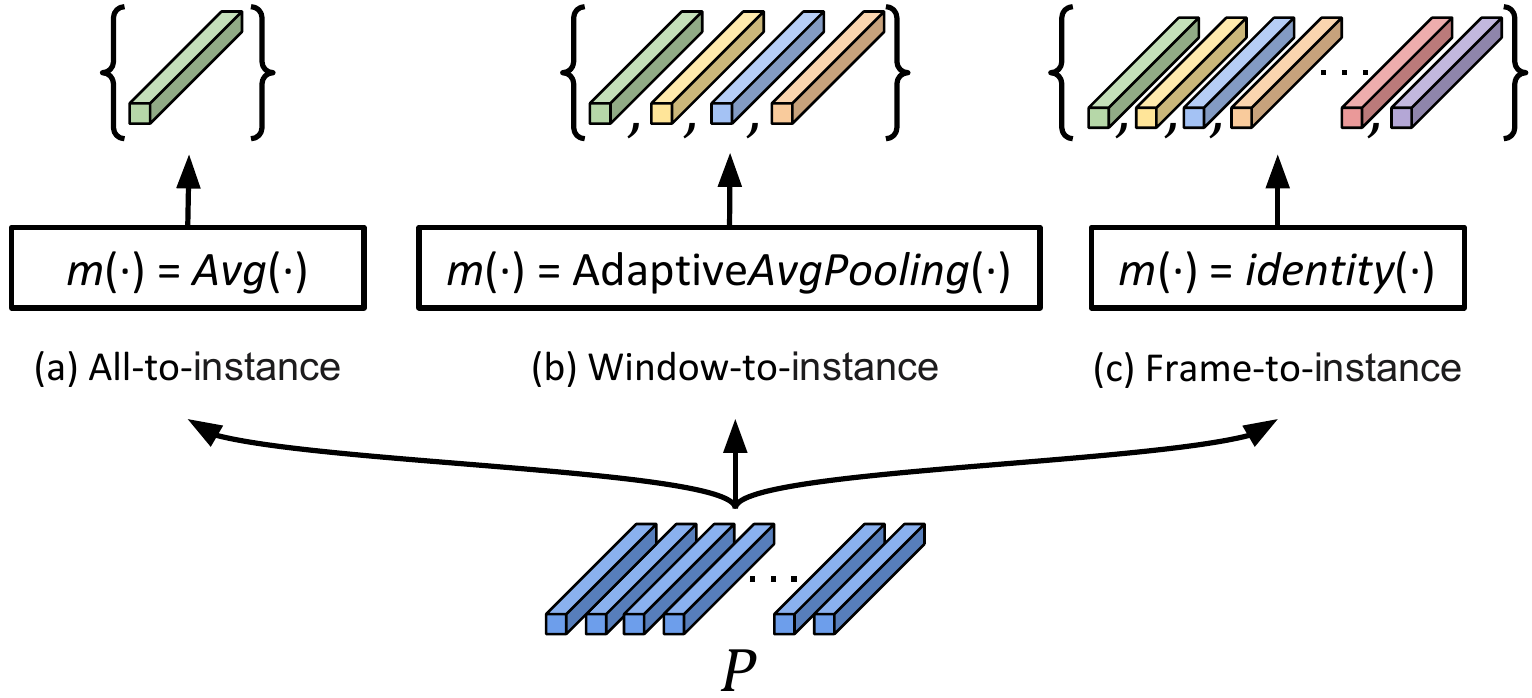}
  \vspace{-0.1cm}
  \caption{\textbf{Instance-mapping function.} This function yields separate instances for the contrastive loss out of each (possibly projected) feature map. The all-to-instance mapping (a) averages all the frames and thus improves robustness to sequence-level misalignment. On the other hand, the frame-to-instance alternative (c) maps each frame to a separate instance, which enlarges the number of negative examples. The window-to-instance mapping (b) represents a trade-off between these options.
    }
  \label{fig:insatnce_mapping_func}
  \vspace{-0.3cm}
\end{figure}

\begin{figure*}[ht!]
\normalsize
  \centering
  \includegraphics[width=0.95\textwidth]{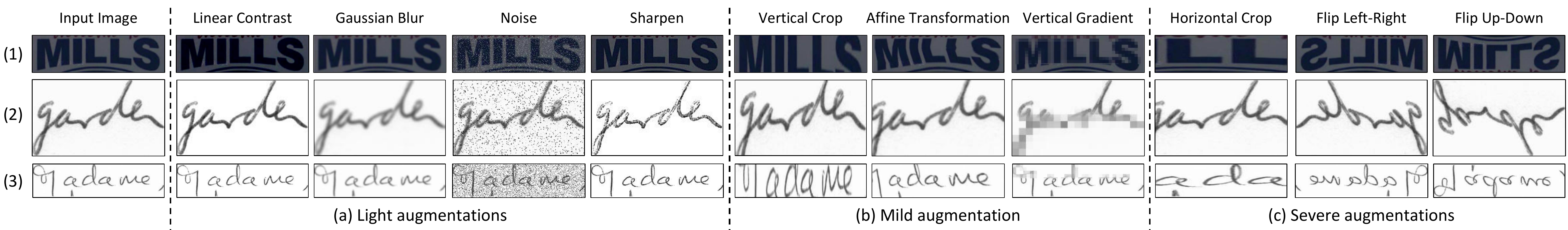}
  \caption{\textbf{Considered augmentations.} Examples of different augmentation types considered in this work, illustrated on three datasets: (1)-IIT5k, (2)-IAM and, (3)-RIMES. As discussed in Section \ref{sec:contrastive_framework}, while flipping and aggressive horizontal cropping are fundamental augmentations in learning visual representations for classification, they should be avoided for text recognition training as they cause sequence-level misalignment which leads to poor contrastive learning.}
  \label{fig:aug_viz}
  \vspace{-0.3cm}
\end{figure*}

%-------------------------------------------------------------------------
\section{Sequence-to-Sequence Contrastive Learning}
\label{sec:contrastive_framework}

Inspired by self-supervised methods for visual representation learning \cite{chen2020simple,falcon2020framework, he2020momentum, chen2020big, caron2020unsupervised}, we propose a contrastive learning framework for sequence-to-sequence visual recognition.
To do so, we introduce a novel instance-mapping stage that yields a separate instance from every few consecutive frames in the sequential feature map.
These instances then serve as the atomic elements in the contrastive loss.
In addition, we design an augmentation procedure that maintains the sequential structure (\cref{fig:aug_receptive_field}) which, as demonstrated in \cref{sec:experiments}, is crucial for yielding effective representations.

As depicted in \cref{fig:main_fig_framework}, we suggest a framework consisting of the following five building-blocks:
\begin{enumerate}
    \item A \emph{stochastic data augmentation} module that is designed to ensure a sequence-level alignment. This operation transforms any given image $\rmX_i$ in a batch of $N$ images, into two augmented images $\rmX_i^a,\rmX_i^b \in \R^{C \times H \times W_i}$, where $C$ denotes the number of input channels, $H$ the image height, and $W_i$ the width of each image which may vary.
    \item A \emph{base encoder} $f(\cdot)$ consisting of several blocks of the recognizer scheme (\cref{fig:main_fig_text_recog}(a)). For each pair of augmented images, this component extracts a pair of sequential representations, $\rmR_i^a, \rmR_i^b \in \R^{F \times T_i}$, where $F$ is the feature dimension, and $T_i$ is the number of frames (columns) which is dependent on the image width (\cref{fig:main_fig_text_recog}(b)).
    \item An optional \emph{projection head} $g(\cdot)$, that transforms the representations using a small auxiliary network, as in~\cite{chen2020simple}. We suggest new projection head types that can handle varying sequence sizes, and denote this stage output by $\rmP_i^a, \rmP_i^b \in \R^{F' \times T_i}$, where $F'$ is the feature dimension after the projection.
    \item A novel \emph{instance-mapping function} $m(\cdot)$ is utilized before the contrastive loss to yield $T'_i$ instances out of $T_i$ projected frames, as illustrated in \cref{fig:insatnce_mapping_func}. 
    These instances are then used as the atomic elements in the contrastive loss. Next, we collect all the instances in the batch into two aligned sets $\Z^a, \Z^b$, each of size $\sum_{i=1}^N T'_i$, such that corresponding indices refer to corresponding frames of the same input image.
    \item A \emph{contrastive loss} function as in \cite{chen2020simple,he2020momentum,falcon2020framework}, that aims to pull closer together representations of corresponding indices of $\Z^a, \Z^b$, i.e. positive pairs, and to push all the others, i.e. negative examples, farther apart:
        \begin{multline}
            \mathcal{L}(\Z^a, \Z^b) = \sum_{r\in \left|\Z^a\right|} \ell_{\text{NCE}} \left(\rvz^a_r, \rvz^b_r; \, \Z^a \cup \Z^b \right) \\
            + \sum_{r\in |\Z^b|} \ell_{\text{NCE}} \left( \rvz^b_r, \rvz^a_r; \, \Z^a \cup \Z^b \right),
        \end{multline}
        where $\ell_{\text{NCE}}(\cdot)$ is the noise contrastive estimation (NCE) loss function~\cite{oord2018representation}:
        \begin{equation*}
            \ell_{\text{NCE}}(\rvu^a, \rvu^b; \U) = -\log \frac{ \exp\left( \simop(\rvu^a, \rvu^b) / \tau \right)}{\sum_{\rvu \in \U\setminus \rvu^a}\exp\left(\simop(\rvu^a, \rvu) / \tau \right)}.
        \end{equation*}
        As in~\cite{chen2020simple}, for the similarity operator we use the cosine distance, $\simop(\rvv, \rvu) = \rvv^T \rvu / \norm{\rvv} \norm{\rvu}$.
\end{enumerate}

We now detail each of these components and describe their configurations.

\paragraph{Data augmentation}
As pointed out in previous papers~\cite{chen2020big, chen2020improved, bachman2019learning}, the augmentation pipeline plays a key part in the final quality of the learned visual representations.
Current stochastic augmentation schemes~\cite{chen2020simple} are mostly based on aggressive cropping, flipping, color distortions, and blurring.
These schemes cannot properly serve the task of text recognition, as they often render the text in the image unreadable.
For example, we should refrain from aggressive horizontal cropping as this might cut out complete characters.

In addition, these augmentation compositions were tailored for tasks as object recognition or classification, where images are atomic input elements in the contrastive loss.
However, since in our framework individual instances are part of a sequence, we design an augmentation procedure that ensures sequence-level alignment.
Therefore, we avoid transformations such as flipping, aggressive rotations, and substantial horizontal translations.
\Cref{fig:aug_viz} depicts different augmentation types considered in this work, including vertical cropping, blurring, random noise, and different perspective transformations.

\paragraph{Base encoder}
The encoder extracts sequential representations from the augmented images $\rmX_i^a, \rmX_i^b$.
While in most contrastive learning schemes the identity of the representation layer is pretty clear -- usually the visual backbone output~\cite{chen2020simple, falcon2020framework}, in text recognition there are different options.
In particular, we consider two candidates as the sequential representations $\rmR_i \in \R^{F \times T_i}$, where each option defines $f(\cdot)$ as the text recognizer scheme (\cref{fig:main_fig_text_recog}) up to this stage:
\begin{enumerate}
    \item The visual features, $\rmR_i = \rmV_i$.
    \item The contextual feature map, $\rmR_i = \rmH_i$, which better captures contextual information within the sequence.
\end{enumerate}

\paragraph{Projection head}
The representations are optionally transformed by a projection head, $\rmP_i = g(\rmR_i)$, which is a small auxiliary neural network that is discarded entirely after the pre-training stage.
As indicated in~\cite{chen2020simple,chen2020big}, this mapping improves the quality of the learned representations.

Currently, a commonly used projection head is the multilayer perceptron (MLP)~\cite{chen2020simple,chen2020big}; however, it can only accommodate fixed-size inputs and thus cannot serve text images.
Therefore, we propose two new projection heads in light of the instance-mapping functions defined below: an MLP projection head that operates on each frame independently as the frame-to-instance mapping (\cref{fig:insatnce_mapping_func}(c)), and a BiLSTM projection head for improving contextual information in the others mappings (\cref{fig:insatnce_mapping_func}(a,b)).

\paragraph{Instance-mapping}
Previous work~\cite{chen2020simple,falcon2020framework,he2020momentum,bachman2019learning} considered images as atomic input elements in the contrastive loss. Therefore, each projected map was vectorized to a single instance, $\rvz_i = \operatorname{flatten}(\rmP_i)$.
However, the inputs and the feature maps in text recognition are of varying sizes and thus cannot be handled by the flatten operator.
% We refer to this operation as a \emph{matrix-to-instance} mapping.
More importantly, in text recognition the feature maps have a sequential structure and thus do not represent a single class. Therefore, we propose to view every few consecutive frames in the feature map as an atomic input element for the contrastive loss.

We propose two approaches for creating individual instances out of sequential feature maps of varying sizes.
In the first approach, we transform every \emph{fixed number of frames} into separate instances, for example, by averaging each $W$ consecutive frames.
In the second approach, we \emph{fix the number of instances} created out of each image, for example, by using adaptive average pooling.

In particular, as depicted in \cref{fig:insatnce_mapping_func}, we consider three instance-mapping functions as specifications of these approaches, which extract $T'_i$ instances out of $T_i$ given frames:
\begin{enumerate}
    \item \textbf{All-to-instance}: All the frames in a sequential feature map are averaged to a single instance, $m(\rmP) = \operatorname{Avg}(\rmP)$, resulting in sets $\Z^a, \Z^b$ of $N$ instances each. 
    % This mapping of a single representation to each input image resembles the common approach in contrastive learning schemes for image classification~\cite{chen2020simple}.
    \item \textbf{Window-to-instance}: Create an instance out of every few consecutive frames. We choose to fix the number of instances and use adaptive average pooling to obtain $T'$ instances. Thus, this operation results in sets $\Z^a, \Z^b$ of size $N \cdot T'$ each.
    \item \textbf{Frame-to-instance}: Each frame is considered as a separate instance, $T'_i = T_i$, resulting in sets of size $\sum_{i=1}^N T_i$, which depend on the input sizes.
\end{enumerate}

Averaging over frames compensates for sequence-level misalignment, which is especially needed for dealing with text written in arbitrary shapes as in scene text images (see \cref{subsec:decoder_evaluation} below).
On the other hand, this operation reduces the number of negative examples in each batch, which, as demonstrated in \cite{chen2020simple,he2020momentum}, can deteriorate the quality of the learned representation.
In this vein, the window-to-instance mapping represents the trade-off between misalignment robustness and sample efficiency.
Note, however, that there are other components in our framework that can also handle this misalignment, such as the BiLSTM projection head and the sequence modeling in the base encoder.

%-------------------------------------------------------------------------
\section{Experiments}
\label{sec:experiments}

% Decoder evaluation table
\begin{table*}
\normalsize
\begin{center}
\footnotesize
\bgroup
\def\arraystretch{1.1}
    \begin{tabular}{|l|c|p{0.42cm}p{0.42cm}|p{0.42cm}p{0.42cm}|p{0.42cm}p{0.42cm}|p{0.42cm}p{0.42cm}|p{0.42cm}p{0.42cm}|p{0.42cm}p{0.42cm}|} 
      \hline
      \multirow{3}{*}{Method} & \multirow{3}{*}{Decoder} & \multicolumn{6}{c|}{Handwritten Dataset} & \multicolumn{6}{c|}{Scene-Text Dataset} \\
       &  & \multicolumn{2}{c}{IAM} & \multicolumn{2}{c}{RIMES} & \multicolumn{2}{c|}{CVL} & \multicolumn{2}{c}{IIT5K} & \multicolumn{2}{c}{IC03} & \multicolumn{2}{c|}{IC13} \\
      & & Acc & ED1 & Acc & ED1 & Acc & ED1 & Acc & ED1 & Acc & ED1 & Acc & ED1 \\
        \hline

        \textit{SimCLR} \cite{chen2020simple} & \multirow{5}{*}{CTC} & 4.0 & 16.0 & 10.0 & 20.3 & 1.8 & 11.1 & 0.3 & 3.1 & 0.0 & 1.0 & 0.3 & 5.0 \\

        \textit{SimCLR Contextual} &  & 6.0 & 17.2 & 13.4 & 25.8 & 7.1 & 17.8 & 1.1 & 4.0 & 2.0 & 2.9 & 1.5 & 6.3 \\

        \cdashline{1-1} \AlgoName{} All-to-instance &  & 34.4 & 60.9 & 59.0 & 80.0 & 55.2 & 73.4 & 20.4 & 42.8 & 24.2 & 49.7 & 24.4 & 51.3 \\

        \AlgoName{} Frame-to-instance &  & 29.4 & 53.1 & 57.5 & 77.5 & 64.3 & 76.0 & 3.0 & 11.4 & 4.6 & 12.3 & 4.9 & 17.5 \\

        \AlgoName{} Window-to-instance &  & \textbf{39.7} & \textbf{63.3} & \textbf{63.8} & \textbf{81.8} & \textbf{66.7} & \textbf{77.0} & \textbf{35.7} & \textbf{62.0} & \textbf{43.6} & \textbf{71.2} & \textbf{43.5} & \textbf{67.9} \\

        \hline

        \hline

        \textit{SimCLR} \cite{chen2020simple} & \multirow{5}{*}{Attention} & 16.0 & 21.2 & 22.0 & 28.3 & 26.7 & 30.6 & 2.4 & 3.6 & 3.7 & 4.3 & 3.1 & 4.9 \\

        \textit{SimCLR Contextual} &  & 17.8 & 23.1 & 34.3 & 40.6 & 34.3 & 38.0 & 3.6 & 5.0 & 4.4 & 5.4 & 3.9 & 6.7 \\

        \cdashline{1-1} \AlgoName{} All-to-instance &  & 51.6 & \textbf{65.0} & 77.9 & 85.8 & 73.1 & 75.3 & 37.3 & 51.8 & 48.0 & 62.2 & 45.8 & 60.4 \\

        \AlgoName{} Frame-to-instance &  & 46.6 & 56.6 & 76.6 & 84.5 & 73.5 & 75.9 & 15.3 & 23.7 & 20.8 & 28.7 & 21.4 & 30.6 \\

        \AlgoName{} Window-to-instance &  & \textbf{51.9} & 63.6 & \textbf{79.5} & \textbf{86.7} & \textbf{74.5} & \textbf{77.1} & \textbf{49.2} & \textbf{68.6} & \textbf{63.9} & \textbf{79.6} & \textbf{59.3} & \textbf{77.1} \\

        \hline
    \end{tabular}
\egroup
\tiny
\caption{\textbf{Representation quality.} Accuracy (Acc) and single edit distance (ED1) of the decoder evaluation -- an analog of the linear evaluation for encoder-decoder networks, in which we train a decoder with labeled data on top of a frozen encoder that was pre-trained on unlabeled images. We compare our \AlgoName{} method of different instance-mapping functions (\cref{fig:insatnce_mapping_func}) with the non-sequential method SimCLR \cite{chen2020simple}. Averaging frames in the feature map, as in all-to-instance and window-to-instance mappings, is especially important in scene-text recognition. \cref{tab:fine_tuning} below presents semi-supervised performance, while \cref{tab:sota_results} shows state-of-the-art results in handwritten datasets.
}
\label{tab:decoder_evaluation}
\vspace{-0.6cm}
\end{center}
\end{table*}

In this section, we experimentally examine our method, comparing its performance with the non-sequential \textit{SimCLR} method \cite{chen2020simple} on several handwritten and scene text datasets.
For this goal, we first consider a decoder evaluation protocol, which is an analog to the linear evaluation procedure (\cite{zhang2016colorful, kolesnikov2019revisiting}) for encoder-decoder based networks.
Then, we test our models in semi-supervised settings in which we fine-tune a pre-trained model with limited amounts of labeled training data.
Finally, we find that when fine-tuned on the entire labeled data, our method achieves state-of-the-art results on standard handwritten datasets.

\noindent
\textbf{Datasets}~ We conduct our experiments on several public datasets of handwritten and scene text recognition.
For handwriting we consider the English datasets IAM~\cite{marti2002iam} and CVL~\cite{kleber2013cvl}, and the French dataset RIMES~\cite{grosicki2009icdar}. For scene text, we train on the synthetic dataset SyntText~\cite{gupta2016synthetic}, and test on three real world datasets: IIT5K~\cite{Mishra2012sj}, IC03~\cite{Lucas2003ic03} and IC13~\cite{Karatzas2013ic13}.
We present samples from each dataset and include more details on the datasets in \cref{app:dtatsets}.

\noindent
\textbf{Metrics}~ To evaluate performance, we adopt the metrics of word-level accuracy (Acc) and word-level accuracy up to a single edit distance (ED1).
For the state-of-the-art comparison in handwriting in \cref{tab:sota_results}, we employ the Character Error Rate (CER) and the Word Error Rate (WER) \cite{sueiras2018offline, zhang2019sequence}.

\noindent
\textbf{Contrastive learning configurations}~
While in \cref{sec:matching_components} we study the effect of modifying each component in our framework, in this section we limit ourselves to the best configuration found for each instance-mapping function (\cref{fig:insatnce_mapping_func}): all-to-instance, frame-to-instance and window-to-instance with $T'=5$.
In all of these schemes, the augmentation pipeline consists of linear contrasting, blurring, sharpening, horizontal cropping and light affine transformations, as further detailed in \cref{app:data_augmentation_pipe}, including examples and pseudo-code.
The base encoder contains a sequential modeling, i.e. $\rmR = \rmH$. Since in such a base encoder the projection head might be redundant (see \cref{sec:matching_components}), we maximize over having and discarding a projection head.
To compare our method to non-sequential contrastive learning, we re-implement the \textit{SimCLR} scheme~\cite{chen2020simple} where the visual features are the representation layer ($\rmR = \rmV$).
For a fair comparison, we consider also \textit{SimCLR Contextual} where the representation layer is the contextual features ($\rmR = \rmH$).

Additional implementation details are described in \cref{app:implementation_details}, including the recognizer settings and the procedures for pre-training, decoder evaluation and fine-tuning.

% Fine-tuning table
\begin{table*}
\normalsize
\begin{center}
\footnotesize
\bgroup
\def\arraystretch{1.1}
    \begin{tabular}{|l|c|ccc|ccc|ccc|ccc|} 
        \hline
        \multirow{4}{*}{Method} & \multirow{4}{*}{Decoder} & \multicolumn{9}{c|}{Handwritten Dataset} & \multicolumn{3}{c|}{Scene-Text Dataset} \\
        & & \multicolumn{3}{c}{IAM} & \multicolumn{3}{c}{RIMES} & \multicolumn{3}{c|}{CVL} & IIT5K & IC03 & IC13\\
        & & \multicolumn{9}{c|}{Label fraction} & \multicolumn{3}{c|}{Label fraction} \\
        & & 5\% & 10\% & 100\% & 5\% & 10\% & 100\% & 5\% & 10\% & 100\% & 100\% & 100\% & 100\%\\
        \hline

        \textit{Supervised Baseline} & \multirow{6}{*}{CTC} & 21.4 & 33.6 & 75.2 & 35.9 & 59.7 & 86.9 & 48.7 & 63.6 & 75.6 & 76.1 & 87.9 & 84.3 \\

        \textit{SimCLR} \cite{chen2020simple} &  & 15.4 & 21.8 & 65.0 & 36.5 & 52.9 & 84.5 & 52.1 & 62.0 & 74.1 & 69.1 & 83.4 & 79.4 \\

        \textit{SimCLR Contextual} &  & 20.4 & 27.8 & 63.7 & 48.6 & 55.6 & 84.4 & 51.8 & 62.3 & 74.1 & 64.5 & 81.7 & 78.1 \\

        \cdashline{1-1} \AlgoName{} All-to-instance &  & 27.5 & 44.8 & \textbf{76.7} & 50.4 & 66.4 & 89.1 & 60.1 & 69.4 & 76.9 & 74.7 & 88.2 & 83.2 \\

        \AlgoName{} Frame-to-instance &  & \textbf{31.2} & \textbf{44.9} & 75.1 & \textbf{61.8} & \textbf{71.9} & \textbf{90.1} & \textbf{66.0} & \textbf{71.0} & \textbf{77.0} & 69.8 & 84.2 & 81.8 \\

        \AlgoName{} Window-to-instance &  & 26.2 & 42.1 & \textbf{76.7} & 56.6 & 62.5 & 89.6 & 61.2 & 69.7 & 76.9 & \textbf{80.9} & \textbf{89.8} & \textbf{86.3} \\

        \hline

        \hline

        \textit{Supervised Baseline} & \multirow{6}{*}{Attention} & 25.7 & 42.5 & 77.8 & 57.0 & 67.7 & 89.3 & 64.0 & 72.1 & 77.2 & \textbf{83.8} & 91.1 & \textbf{88.1} \\

        \textit{SimCLR} \cite{chen2020simple} &  & 22.7 & 32.2 & 70.7 & 49.9 & 60.9 & 87.8 & 59.0 & 65.6 & 75.7 & 77.8 & 88.8 & 84.9 \\

        \textit{SimCLR Contextual} &  & 24.6 & 32.9 & 70.2 & 51.9 & 63.0 & 87.3 & 59.7 & 66.2 & 75.2 & 72.2 & 87.0 & 82.3 \\

        \cdashline{1-1} \AlgoName{} All-to-instance &  & \textbf{40.3} & 51.6 & 79.8 & 69.7 & 76.9 & \textbf{92.5} & 69.5 & 73.2 & 77.6 & 80.9 & 90.0 & 87.0 \\

        \AlgoName{} Frame-to-instance &  & 37.2 & 48.5 & 78.2 & 68.8 & 75.9 & 92.3 & 69.7 & 73.4 & 77.5 & 76.3 & 90.2 & 85.8 \\

        \AlgoName{} Window-to-instance &  & 38.1 & \textbf{52.3} & \textbf{79.9} & \textbf{70.9} & \textbf{77.0} & 92.4 & \textbf{73.1} & \textbf{74.8} & \textbf{77.8} & 82.9 & \textbf{92.2} & 87.9 \\

        \hline
    \end{tabular}
\egroup
\tiny
\caption{\textbf{Semi-supervised results.} Accuracy of fine-tuning a pre-trained model with 5\%, 10\% and 100\% of the labeled data. For scene-text datasets we test only for 100\% as the data is anyhow synthetic. As presented in \cref{tab:sota_results}, our method achieves state-of-the-art results on handwritten datasets.}
\label{tab:fine_tuning}
\vspace{-0.3cm}
\end{center}
\end{table*}
% SOTA table
\begin{table}
\normalsize
\begin{center}
\footnotesize
\bgroup
\def\arraystretch{1.1}
    \begin{tabular}{|c|c|ccc|} 
      \hline
      Dataset & Method & WER & CER & Average \\
      \hline
      \multirow{6}{*}{IAM} &
      Bluche et al.~\cite{bluche2015deep} & 24.7 & \textbf{7.3} & 16.00 \\
      & Bluche et al.~\cite{bluche2016joint} & 24.6 & 7.9 & 16.25 \\
      & Sueiras et al.~\cite{sueiras2018offline} & 23.8 & 8.8 & 16.30 \\
      & ScrabbleGAN~\cite{fogel2020scrabblegan} & 23.6 & - & - \\
      & SSDAN*~\cite{zhang2019sequence} & 22.2 & 8.5 & 15.35 \\
      \cdashline{2-5}
      & \AlgoName & \textbf{20.1} & 9.5 & \textbf{14.80} \\
    %   \AlgoName & Test & Train & ? & ? & ?\\
      \hline
      \multirow{4}{*}{RIMES} & Alonso et al.~\cite{alonso2019adversarial} & 11.9 & 4.0 & 7.95 \\
      & ScrabbleGAN~\cite{fogel2020scrabblegan} & 11.3 & - & - \\
      &Chowdhury et al.~\cite{chowdhury2018efficient} & 9.6 & 3.4 & 6.55 \\
      \cdashline{2-5}
      &\AlgoName & \textbf{7.6} & \textbf{2.6} & \textbf{5.5} \\
      \hline
    \end{tabular}
\egroup
\tiny
\caption{\textbf{SOTA error rates.} Word and character error rates (WER and CER) of our method compared to current state-of-the-art word-level methods on IAM and RIMES datasets. '*' indicates using the unlabeled test set for training.}
\label{tab:sota_results}
\vspace{-0.3cm}
\end{center}
\end{table}

\subsection{Decoder evaluation}
\label{subsec:decoder_evaluation}

We start our experimental study by evaluating the quality of the learned visual representation.
To this end, we establish a decoder evaluation protocol that extends the widely-used linear evaluation protocol~\cite{zhang2016colorful, kolesnikov2019revisiting} to encoder-decoder based networks.
In this protocol, we first train a base-encoder $f(\cdot)$ on the unlabeled data, using some self-supervised method.
Then, we freeze the encoder weights and train on top of it a CTC or an attention decoder (\cref{fig:main_fig_text_recog}(c,d)) with all the labeled data.
Since we keep the encoder untouched, this test can be seen as a proxy to the representation learning efficiency.

\Cref{tab:decoder_evaluation} shows the results of our proposed \AlgoName{} method, compared with vanilla \textit{SimCLR}~\cite{chen2020simple} and \textit{SimCLR Contextual}, over public datasets of handwritten and scene text benchmarks, with either a CTC or an attention decoder (\cref{fig:main_fig_text_recog}(c,d)).
As discussed above, current contrastive methods for visual representations are designed for tasks such as classification and object detection, where images are atomic input elements.
However, in text recognition, a word is viewed as a sequence of characters, and therefore, the standard 'whole image' concept leads to poor performance.
Specifically, the augmentation procedure considered in~\cite{chen2020simple,he2020momentum} usually breaks the sequential structure of the input text image. In addition, in these prior papers, the feature map is treated as a single representation, whereas in text recognition, it is eventually decoded as a sequence of representations.

The comparison between the different instance-mapping functions demonstrates that the best results are achieved by the window-to-instance mapping (\cref{fig:insatnce_mapping_func}(b)).
As can be seen, the frame-to-instance mapping, which does not average consecutive frames, performs poorly on scene text images. These images are prone to sequence-level misalignment by even mild augmentations, as they contain text that already comes in diverse shapes.
On the other hand, the all-to-instance mapping, which averages all the frames, significantly reduces the number of negative examples in each batch, which in turn, also affects performance.
The window-to-instance mapping succeeds in balancing these concerns and therefore leads to better performance.

\subsection{Fine-tuning}
\label{subsec:finetuning}
We further evaluate our method by considering semi-supervised settings. We use the same encoders as before, which were pre-trained on the unlabeled data, but now let the whole network be fine-tuned using 5\% or 10\% of the labeled dataset. 
Contrary to prior work~\cite{zhai2019s4l,chen2020simple}, which consider class-balanced datasets, we simply use the same randomly selected data for all the experiments.
We also test for fine-tuning on the entire labeled data, as suggested in \cite{chen2020simple}.
Note that this is the only evaluation we examine for scene text recognition, as the training dataset is anyhow synthetic in this case.

As opposed to the decoder evaluation, here, the goal is to achieve the best results and not just qualify the learned representations. Therefore, contrary to the decoder evaluation, in the fine-tuning phase, one can attach additional layers besides the decoder on top of the encoder. That said, we only attach a text decoder (CTC or attention), as the base encoder in the following experiments already contains a sequence modeling.

\cref{tab:fine_tuning} compares our method with \textit{SimCLR} \cite{chen2020simple}, \textit{SimCLR Contextual}~and \textit{supervised baseline} training. 
As can be seen, in the case of text recognition, pre-training using non-sequential contrastive learning schemes often leads to deterioration in performance compared to the \textit{supervised baseline}.
\AlgoName{}, on the other hand, achieves better performance for every semi-supervised scenario and on every handwritten dataset.
In particular, the window-to-instance mapping performs the best for the attention decoder, while the frame-to-instance alternative is superior when using the CTC decoder. This is an interesting result that might indicate that frame-to-instance better fits the CTC decoder as they both operate on individual frames of the feature map.

In the case of fine-tuning on 100\% of the labeled data, although our method does not use any additional data, it still succeeds in significantly improving the results of the fully \textit{supervised baseline} training on handwritten datasets. In particular, our method gains an improvement of +1.9\% on average for the CTC decoder and +1.7\% on average for the attention decoder. 
In scene text datasets, \AlgoName{} achieves an improvement of 2.9\% on average for the CTC decoder; however, it performs the same on average as the \textit{supervised baseline} for the attention decoder.
The mixed performance in scene text might be a result of utilizing only synthetic data in the representation learning phase.

Notably, as presented in \cref{tab:sota_results}, \AlgoName{} using window-to-instance mapping outperforms the current word-level state-of-the-art performance on both IAM and RIMES datasets, even when compared to methods which used the test-set in their unsupervised training.
Note that for a fair comparison, we only include results that considered the same test-set and that did not attach a language model.

\begin{figure}[t!]
    \centering
    \begin{subfigure}[t]{0.24\textwidth}
        \centering
        \includegraphics[width=\linewidth]{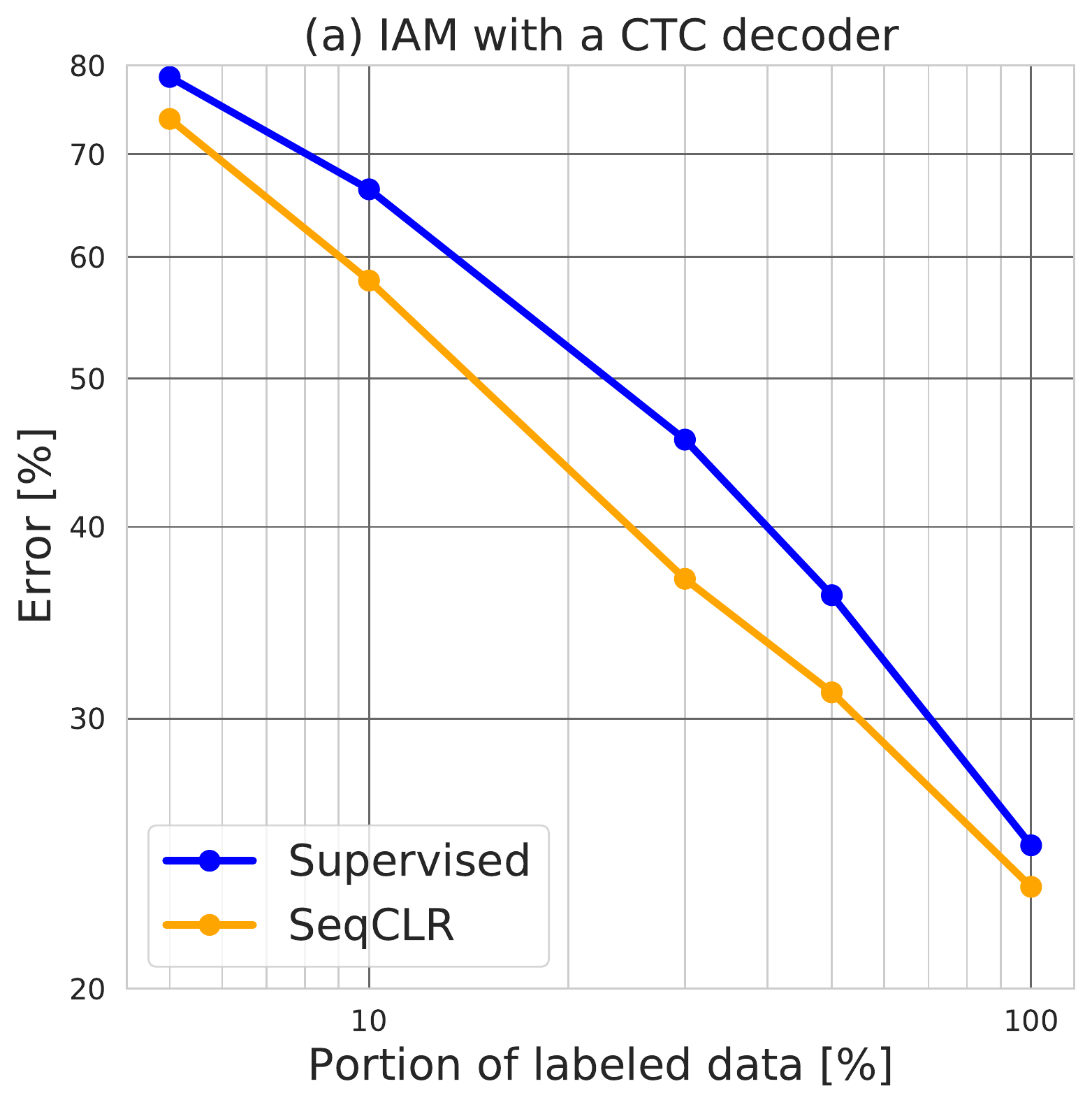}
    \end{subfigure}%
    ~ 
    \begin{subfigure}[t]{0.24\textwidth}
        \centering
        \includegraphics[width=\linewidth]{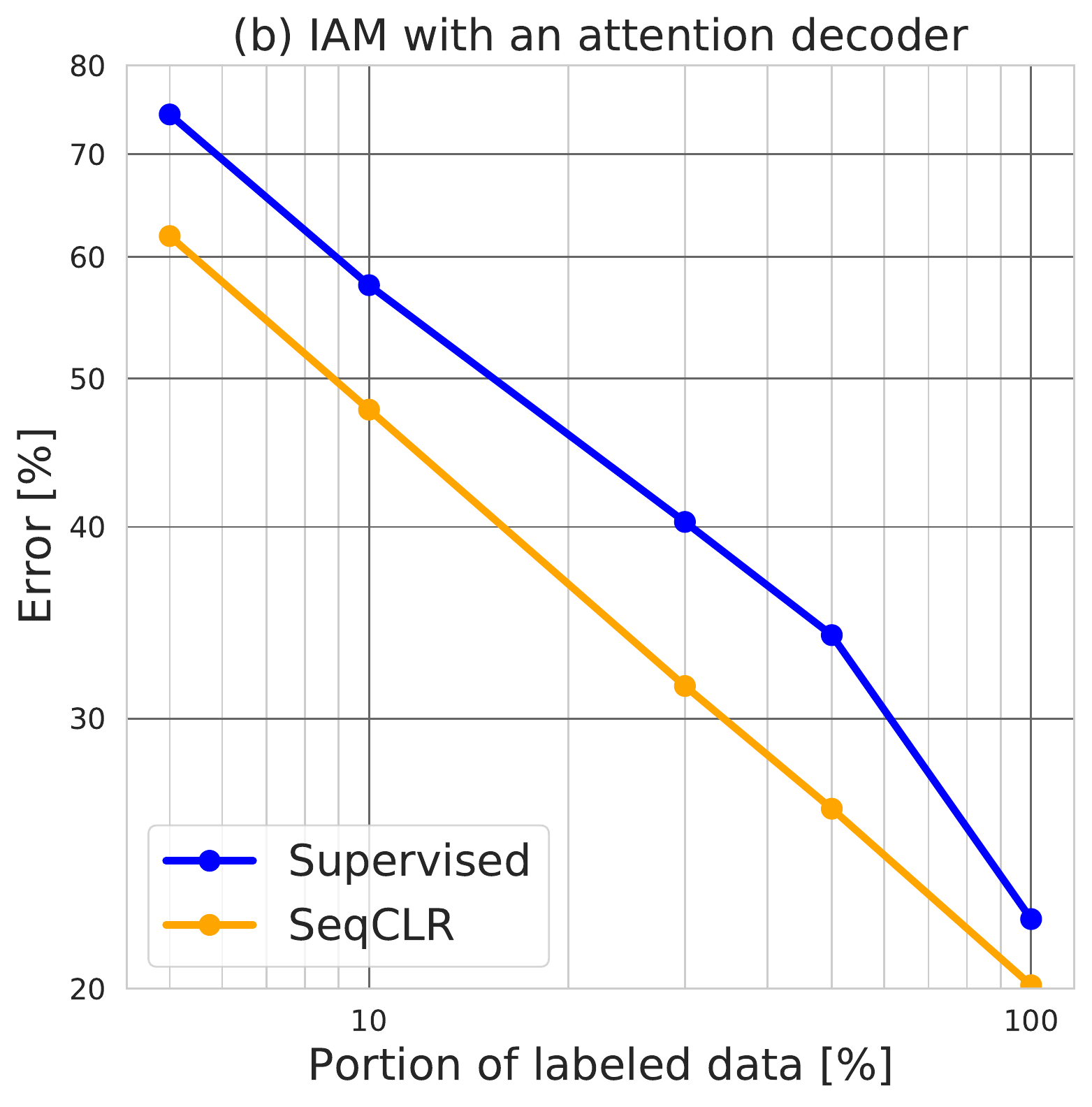}
    \end{subfigure}
    
    \begin{subfigure}[t]{0.24\textwidth}
        \centering
        \includegraphics[width=\linewidth]{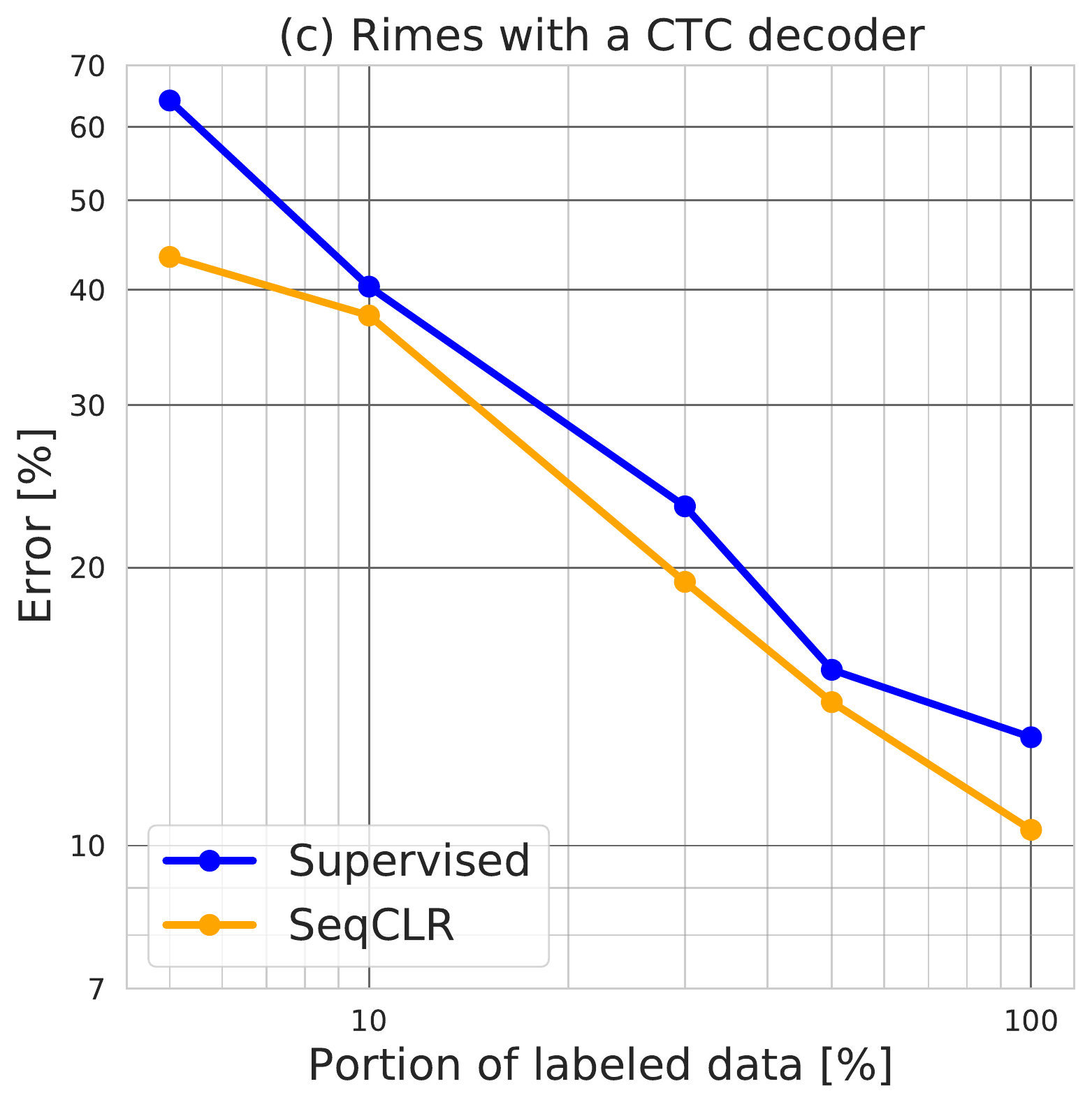}
    \end{subfigure}%
    ~ 
    \begin{subfigure}[t]{0.24\textwidth}
        \centering
        \includegraphics[width=\linewidth]{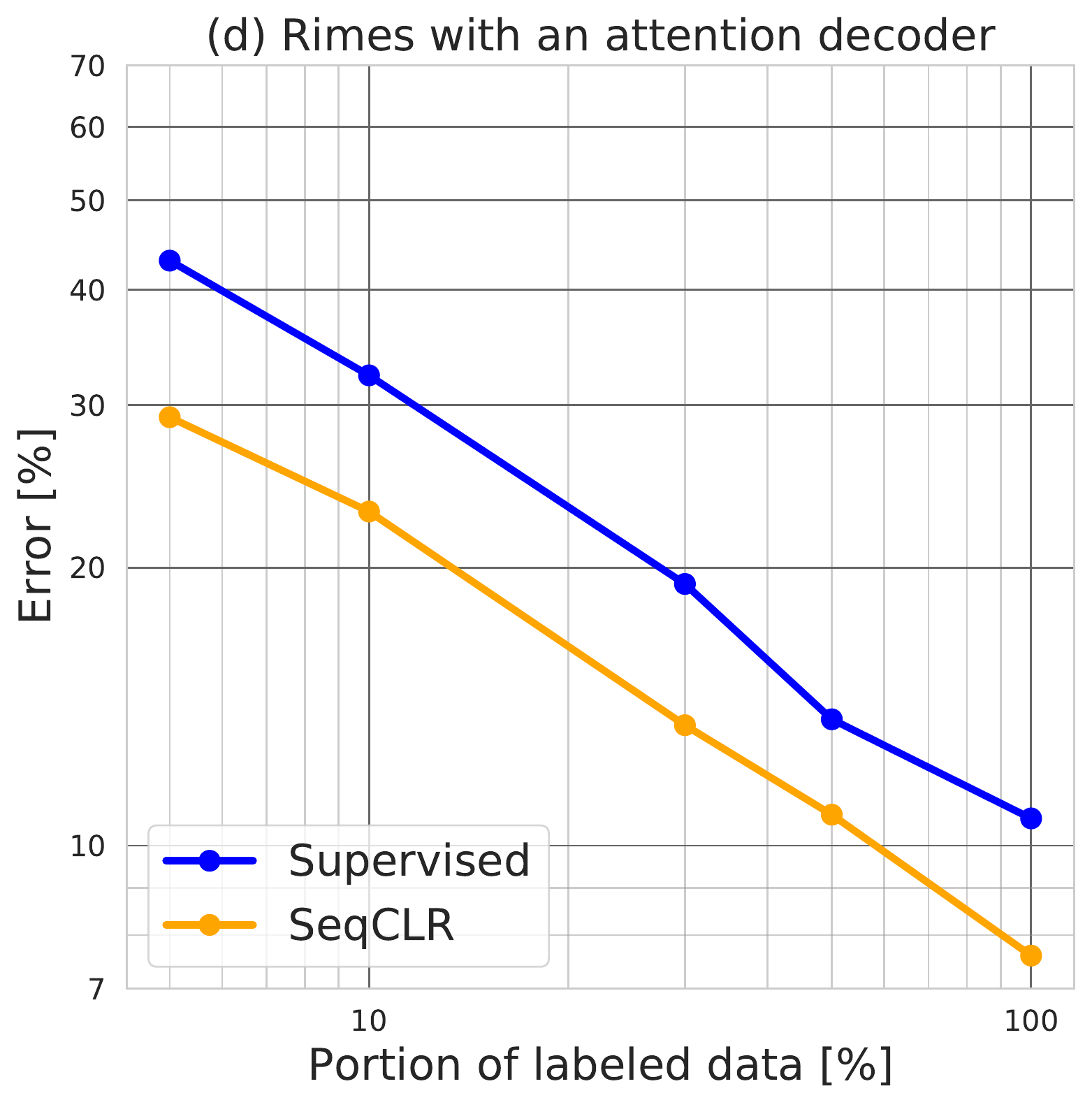}
    \end{subfigure}
    \caption{\textbf{Word error rate as a function of labeled data amount in a log-log scale.} Roughly speaking, \AlgoName{} unsupervised pre-training has the same effect as doubling the labeled data amount in the sense of reducing the error rate.}
    \label{fig:loglog}
\end{figure}

In \cref{fig:loglog}, we present the word error rate of \AlgoName{} and the \textit{supervised baseline} as a function of the portion of the labeled data for IAM and RIMES datasets using CTC and attention decoders. As can be seen, in the RIMES dataset using an attention decoder, \AlgoName{} utilizing $50\%$ of the labeled data achieves the same error rate as the current state-of-the-art algorithm trained on the entire labeled data.

%-------------------------------------------------------------------------
\section{Ablation Study}
\label{sec:matching_components}
% Projection-study table
\begin{table}
\normalsize
\begin{center}
\scriptsize
\bgroup
\def\arraystretch{1.1}
    \begin{tabular}{|l|l|c|c|c|c|} 
      \hline
      \multirow{2}{*}{Projection head} & \multirow{2}{*}{Mapping $m(\cdot)$} & \multicolumn{2}{c|}{Visual feat.} & \multicolumn{2}{c|}{Contextual feat.} \\
      & & CTC & Attn & CTC & Attn \\ 
        \hline

        None & \multirow{3}{*}{All-to-instance} & 2.0 & 55.8 & 55.4 & \textbf{77.9} \\

        MLP per frame &  & 29.9 & 70.3 & 50.4 & 72.5 \\

        BiLSTM &  & \textbf{35.5} & \textbf{70.9} & \textbf{59.0} & 74.8 \\

        \hline None & \multirow{3}{*}{Frame-to-instance} & 27.4 & 56.9 & 49.9 & 75.8 \\

        MLP per frame &    & \textbf{39.9} & 69.8 & \textbf{57.5} & \textbf{76.6} \\

        BiLSTM &    & 37.4 & \textbf{69.9} & 43.5 & 64.3 \\

        \hline None & \multirow{3}{*}{Window-to-instance} & 27.9 & 67.6 & 59.9 & \textbf{79.5} \\

        MLP per frame &   & 29.9 & 70.3 & 50.4 & 72.5 \\

        BiLSTM &   & \textbf{35.8} & \textbf{74.0} & \textbf{63.8} & 75.9 \\
      \hline
    \end{tabular}
\egroup
\scriptsize
\caption{\textbf{Matching projection heads to mappings}. Representation qualities (decoder evaluation accuracy) of combining different projection heads with instance-mapping functions (\cref{fig:insatnce_mapping_func}).
While BiLSTM head fits the all-to-instance and window-to-instance mappings, the MLP per frame performs better with the frame-to-instance mapping.
% While MLP per frame works better the frame-to-instance mapping, BiLSTM works best with all-to-instance and window-to-instance functions.
% on the RIMES dataset trained with different configuration of our contrastive framework, particularly the base encoder $f(\cdot)$, the mapping function $m(\cdot)$ and the projection head type.
}
\label{tab:projection_ablation}
\vspace{-0.3cm}
\end{center}
\end{table}

\begin{figure}[t]
  \centering
  \includegraphics[width=\columnwidth]{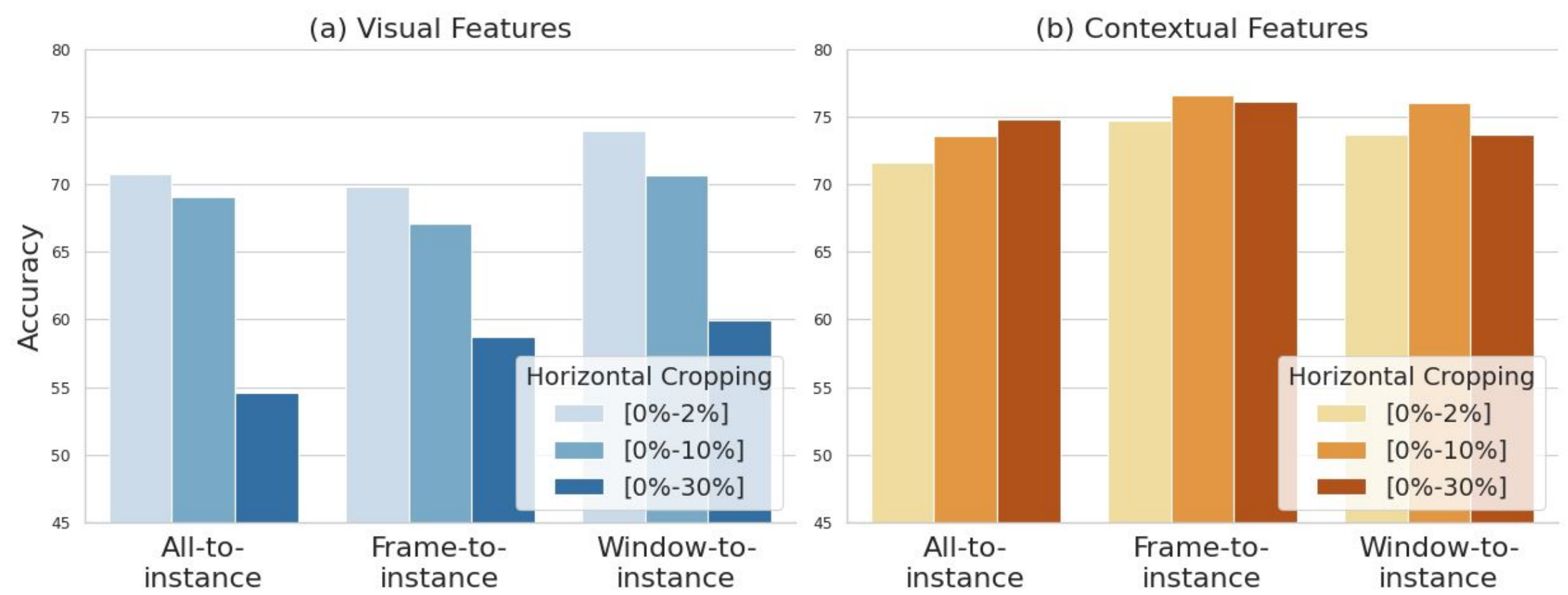}
  \caption{\textbf{Horizontal cropping.} The effect of horizontal cropping on the representation learning. This augmentation affects learning when the representation layer is chosen to be the visual features (a); however, a light version of it can help in encoders that contain sequence modeling (b).
}
  \label{fig:horizontal_cropping_effect}
\end{figure}

After studying the effect of the instance-mapping functions in \cref{subsec:decoder_evaluation}, in this section, our goal is two-fold. First, to match the components in our framework, and second, to demonstrate the importance of maintaining sequence-level alignment by applying different augmentation procedures.
% We refer the reader to \cref{app:experiments} for further experimental study.

For evaluating the representation learning in this section, we adopt the decoder evaluation protocol (\cref{subsec:decoder_evaluation}) on the RIMES dataset, considering representation layers of visual features ($\rmR = \rmV$) and contextual features ($\rmR = \rmH$).

\paragraph{Matching the components}
As can be seen in \cref{tab:projection_ablation}, including a sequence modeling in the base encoder ($\rmR = \rmH$) leads to significant improvements in the quality of the learned representation for both CTC and attention decoders.
In general, incorporating a projection head also improves representation effectiveness; however, when utilizing a sequence modeling and an attention decoder, each containing BiLSTM layers, then the BiLSTM projection head appears as a redundant component.
The key message from these experiments is that the \AlgoName{} components should be selected dependently.

\paragraph{Sequence-level alignment}

In text recognition, individual instances are part of a sequence, and thus, the augmentation procedure needs to maintain a sequence-level alignment (\cref{fig:aug_receptive_field}).
On the other hand, as suggested in~\cite{chen2020simple,chen2020improved}, strong data augmentations contribute to contrastive learning.
The following experiments aim to study this trade-off and to identify components in our framework that improve the robustness to sequence-level misalignment.

\Cref{fig:horizontal_cropping_effect} demonstrates the effect of horizontal cropping on the representation learning. As observed, having a base encoder with sequence modeling is crucial for handling sequence-level misalignment caused by even mild horizontal cropping.
Note that such cropping might cut out complete characters from the input text images.
This indicates that during the representation learning, the sequence modeling successfully captures contextual information within the visual features, which compensates for missing visual information and sequence-level misalignment.

%-------------------------------------------------------------------------
\section{Discussion and conclusions}
\label{sec:conclusion}

We presented \AlgoName{}, a contrastive learning algorithm for self-supervised learning of sequence-to-sequence visual recognition that divides each feature map into a sequence of individual elements for the contrastive loss.
In order to take full-advantage of self-supervision, we proposed a number of sequence-specific augmentation techniques that differ from whole-image equivalents.

The main take-home lesson is that paying attention to the task's structure, i.e.~treating an image as a sequence of frames, pays off. Our experiments show that \AlgoName{} largely outperforms current non-sequential contrastive learning methods in recognizing handwritten and scene text images when the amount of supervised training is limited. Furthermore, our method achieves state-of-the-art performance on handwriting -- compared with the best methods in the literature  \AlgoName{} reduces the word error rate by 9.5\% and 20.8\% on IAM and RIMES, the standard benchmark datasets.
\AlgoName{} is the result of careful experimental evaluation of different design options, including different augmentation compositions, encoder architectures, projection heads, instance-mapping functions, and decoder types. 
% We find that these components should be selected dependently. 
% For example, strong data augmentations, for the augmentation procedure or inherently as in scene text images, requires a contextual modeling in the base-encoder and an instance-mapping function of window-to-instance or all-to-instance.

The success of \AlgoName{} will hopefully encourage other researchers to explore semi-supervised and self-supervised schemes for text recognition, as well as contrastive learning algorithms for different sequence-to-sequence predictions.

%-------------------------------------------------------------------------

\newpage

{\small
\bibliographystyle{ieee_fullname}
\bibliography{bib}
}

\clearpage

\appendix

\section{Text Recognition Scheme}
\label{app:text_recog_scheme}
In this section, we provide additional details on the text recognition architecture components considered in this work. In particular, we focus on three components: (i) the transformation performed by the thin-plate splines (TPS)~\cite{shi2016robust, jaderberg2015spatial}, (ii) the CTC based decoder~\cite{Graves2006ctc, Bai2015crnn} and (iii) the attention based decoder~\cite{baek2019wrong, bai2017accurate, litman2020scatter}.

\subsection{Transformation}
This stage transforms a cropped text image $\rmX$ into a normalized image $\rmX^\prime$.
This step is necessary when the input image contains text in a non-axis aligned layout, as often occurs in handwritten text and scene text images.

\begin{figure}[h!]
  \centering
  \includegraphics[width=\columnwidth]{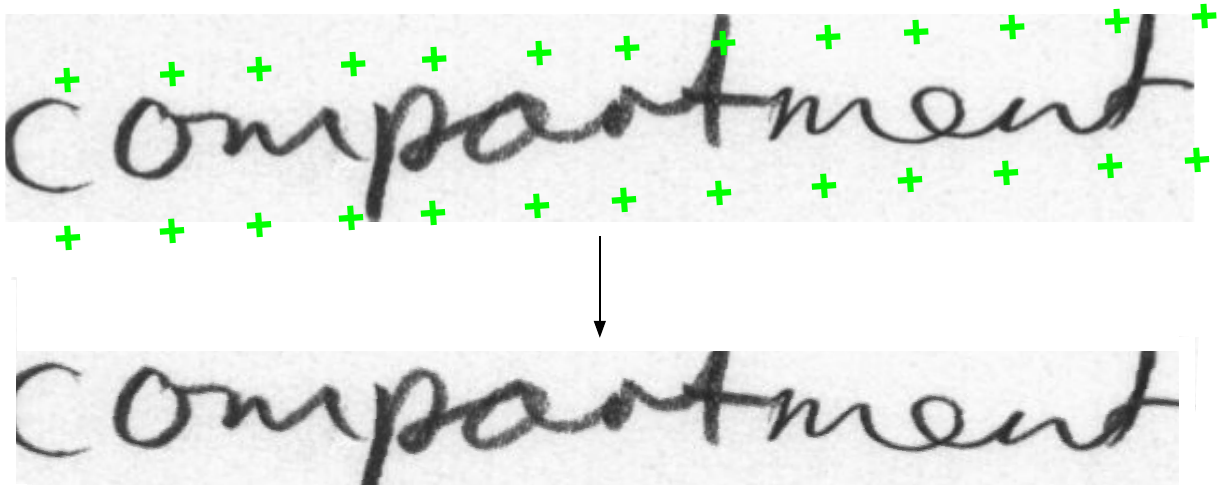}
  \caption{\textbf{TPS transformation.}
  This transformation first predicts fiducial points -- marked as green points.
  Then, a smooth spline interpolation is employed, transforming these points to the border of a constant rectangle, yielding a normalized image with a fixed predefined size.}
  \label{fig:tps_fig}
\end{figure}

In this work, we follow~\cite{baek2019wrong}, and utilize the Thin Plate Spline (TPS) transformation~\cite{shi2016robust, jaderberg2015spatial} which is a variant of the spatial transformer network~\cite{jaderberg2015spatial}.
As depicted in \cref{fig:tps_fig}, in this transformation, we first detect a pre-defined number of fiducial points at the top and bottom of the text region. Then, we apply a smooth spline interpolation between the obtained points to map the predicted textual region to a constant pre-defined size.

\subsection{Connectionist Temporal Classification (CTC)}

The CTC decoder~\cite{Graves2006ctc} operates on a given sequential feature map $\rmF = [\rvf_1,\rvf_2,\ldots,\rvf_T]$, which in our framework can be $\rmF \in \{\rmV, \rmH, (\rmH,\rmV)\}$. The inference phase consists of three stages. 
In the first stage, each frame $\rvf_t$ is transformed by a fully connected layer to yield $\rvf'_t$. Then, the CTC finds the sequence of characters with the highest probability:
\begin{equation}
    \rvc = \argmax_{\rvpi} \prod_{t=1}^T f'_{t, \pi_t},
\end{equation}
where $f'_{t, i}$ denotes the $i$th element in $\rvf'_t$.
Next, the CTC removes repeated characters and blanks:
\begin{equation}
    \rvy = \varphi( \rvc ),
\end{equation}
where $\varphi(\cdot)$ denotes the mapping function. For example, if $\rvc = ``\texttt{aa-a-bbb-cc-ccc-\--}\text{''}$ then ``$\rvy =  \texttt{aabcc}$\text{''}.

For the CTC procedure during training we refer the reader to \cite{Graves2006ctc,Bai2015crnn}.

\subsection{Attention decoder}
As for the CTC, the attention also operates on a given sequential feature map $\rmF \in \{\rmV, \rmH, (\rmH,\rmV)\}$.
The first step of decoding starts by computing the vector of attentional weights, $\rvalpha_{t'} \in \R^T$. For this goal, we first calculate $e_{t',t}$:
\begin{equation}
    \label{equ:equ2}
    e_{t',t} = \rva^T \tanh(\rmW \rvs_{t'-1} + \rmV \rvf_t + \rvb) \,,
\end{equation}
where $\rmW, \rmV, \rva, \rvb$ are trainable parameters, and $\rvs_{t'-1}$ is the hidden state of the recurrent cell within the decoder at time $t'$.
Then, we compute $\rvalpha_{t'}$ by:
\begin{equation}
    \alpha_{t',t} = \frac{\exp (e_{t',t})}{\sum_{j=1}^{T}e_{t',j}}   \,.
\end{equation}
As mention in the paper, the decoder linearly combines the columns of $\rmF$ into a vector $\rvg$ by utilizing the learned $\rvalpha_{t'}$:
\begin{equation}
    \rvg_{t'} = \sum_{t=1}^{T}\alpha_{t',t} \rvf_t \,.
\end{equation}
Next, the recurrent cell is fed with:
\begin{equation}
    (\rvx_{t'},\rvs_{t'}) = \operatorname{RNN}\left(\rvs_{t'-1}, \left[\rvg_{t'}, f(\rvy_{t'-1})\right]\right) \,,
\end{equation}
where $f(\cdot)$ is a one-hot embedding, $[\cdot, \cdot]$ denotes the concatenation operator, and $\rvy_{t'}$ is obtained by:
\begin{equation}
    \rvy_{t'} = \softmax (\rmW_0 \rvx_{t'} + \rvb_0) \,,
\end{equation}
where $\rmW_0, \rvb_0$ are trainable parameters. 
The loss used for the attention decoder is the negative log-likelihood, as in~\cite{bai2017accurate}.

\section{Data Augmentation}
\label{app:data_augmentation_pipe}

\begin{figure}[t!]
  \centering
  \includegraphics[width=\columnwidth]{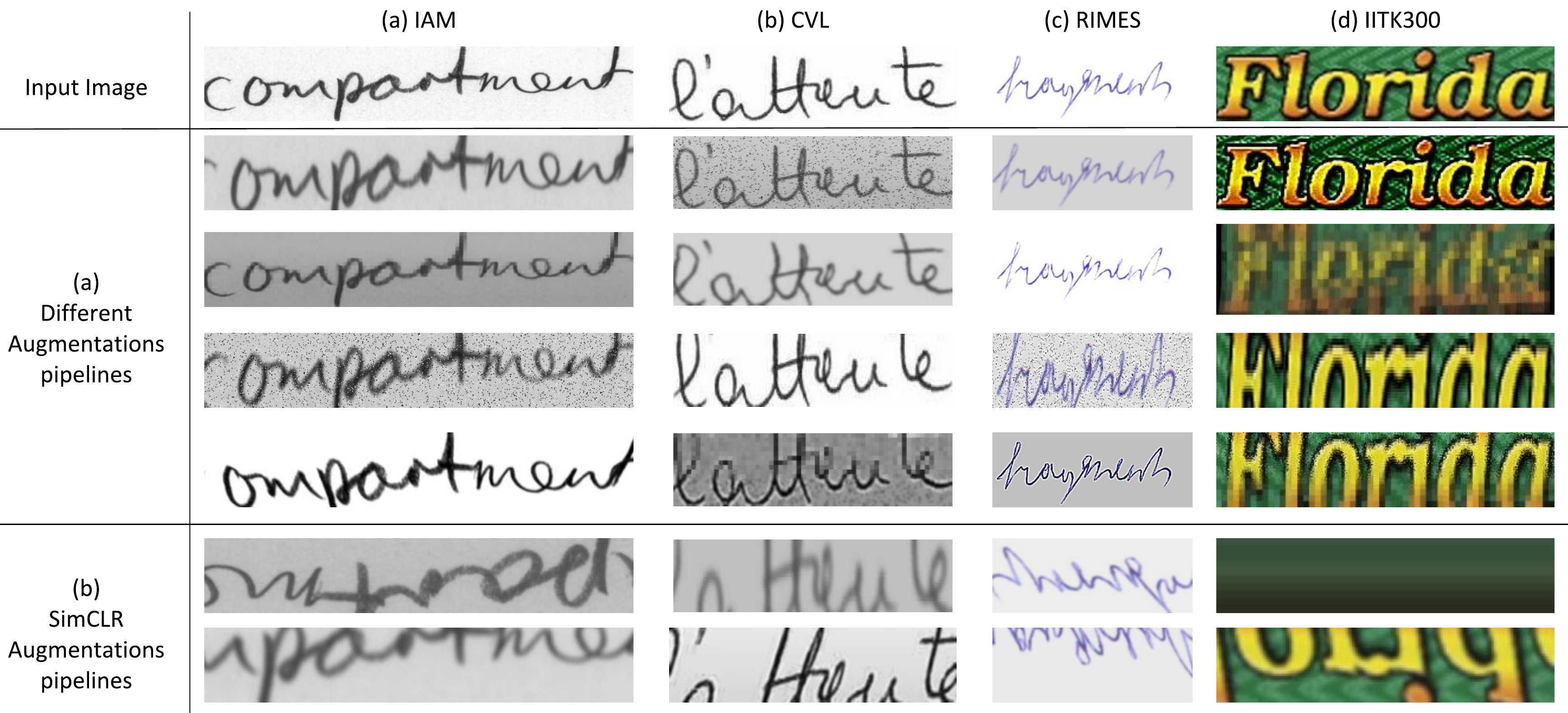}
  \caption{\textbf{Illustrations of augmentation procedures.} We show different augmentation pipelines that were considered in this work (a), which led to the final augmentation pipeline.
  We also show augmentations examples using the SimCLR~\cite{chen2020simple} pipeline (b).}
  \label{fig:aug_pipe_app}
\end{figure}

In this section, we provide additional details for reproducing our augmentation procedure, implemented using the \texttt{imgaug}~\cite{imgaug} augmentation package.
As described in Section 5, this augmentation pipeline is used for the self-supervised training, where we stochastically augment each image twice.
In \cref{fig:aug_pipe_app}, we present different augmentation procedures that we examined in our work, which eventually led us to the final pipeline.
Our default procedure consists of a random subset of the following operations.

\paragraph{Linear contrast} We modify the input image contrast by applying the pixel-wise transformation: $127 + \alpha(v-127)$, where $\alpha$ is sampled uniformly from the interval $[0.5, 1.0]$ and $v\in[0,255]$ is the pixel value.

\paragraph{Blur} We blur the image using a Gaussian kernel with a randomly selected standard deviation of $\sigma \in (0.5, 1.0)$.

\paragraph{Sharpen} The image is sharpened by blending it with a highly sharpened version of itself.
The lightness parameter found in the \texttt{imgaug} framework, is sampled uniformly from the interval $[0.0, 0.5]$, and the alpha factor used for blending the image is sampled uniformly from the interval $[0.0, 0.5]$.

\paragraph{Crop} We first extract a smaller-sized sub-image from the given full-sized input image. Then, we resize this crop to the original size.
As mention in Section 6, the vertical cropping can be more aggressive than the horizontal cropping.
Therefore, the percentage of the vertical cropping is sampled uniformly from the interval $[0\%, 40\%]$, while the horizontal cropping percentage is sampled from $[0\%, 2\%]$. 

\paragraph{Perspective transform} A four point perspective transformation is applied. These points are placed on the input image by using a random distance from the original image corners, where the random distance is drawn from a normal distribution with a standard deviation sampled uniformly from the interval $[0.01, 0.02]$.

\paragraph{Piecewise affine} We apply an affine transformation that moves around each grid point by a random percentage drawn uniformly from the interval $[2\%, 3\%]$.

A pseudo-code for the augmentation pipeline, written with the \texttt{imgaug}~\cite{imgaug} package, is as follows.
\begin{lstlisting}[language=Python]
from imgaug import augmenters as iaa
iaa.Sequential([iaa.SomeOf((1, 5),
[
  iaa.LinearContrast((0.5, 1.0)),
  iaa.GaussianBlur((0.5, 1.5)),
  iaa.Crop(percent=((0, 0.4),
                   (0, 0),
                   (0, 0.4),
                   (0, 0.0)),
                   keep_size=True),
  iaa.Crop(percent=((0, 0.0),
                  (0, 0.02),
                  (0, 0),
                  (0, 0.02)),
                  keep_size=True),
  iaa.Sharpen(alpha=(0.0, 0.5),
            lightness=(0.0, 0.5)),
  iaa.PiecewiseAffine(scale=(0.02, 0.03),
                    mode='edge'),
  iaa.PerspectiveTransform(
                    scale=(0.01, 0.02)),
],
random_order=True)])
\end{lstlisting}

\section{Datasets}
\label{app:dtatsets}

\begin{figure}[t]
  \centering
  \includegraphics[width=\columnwidth]{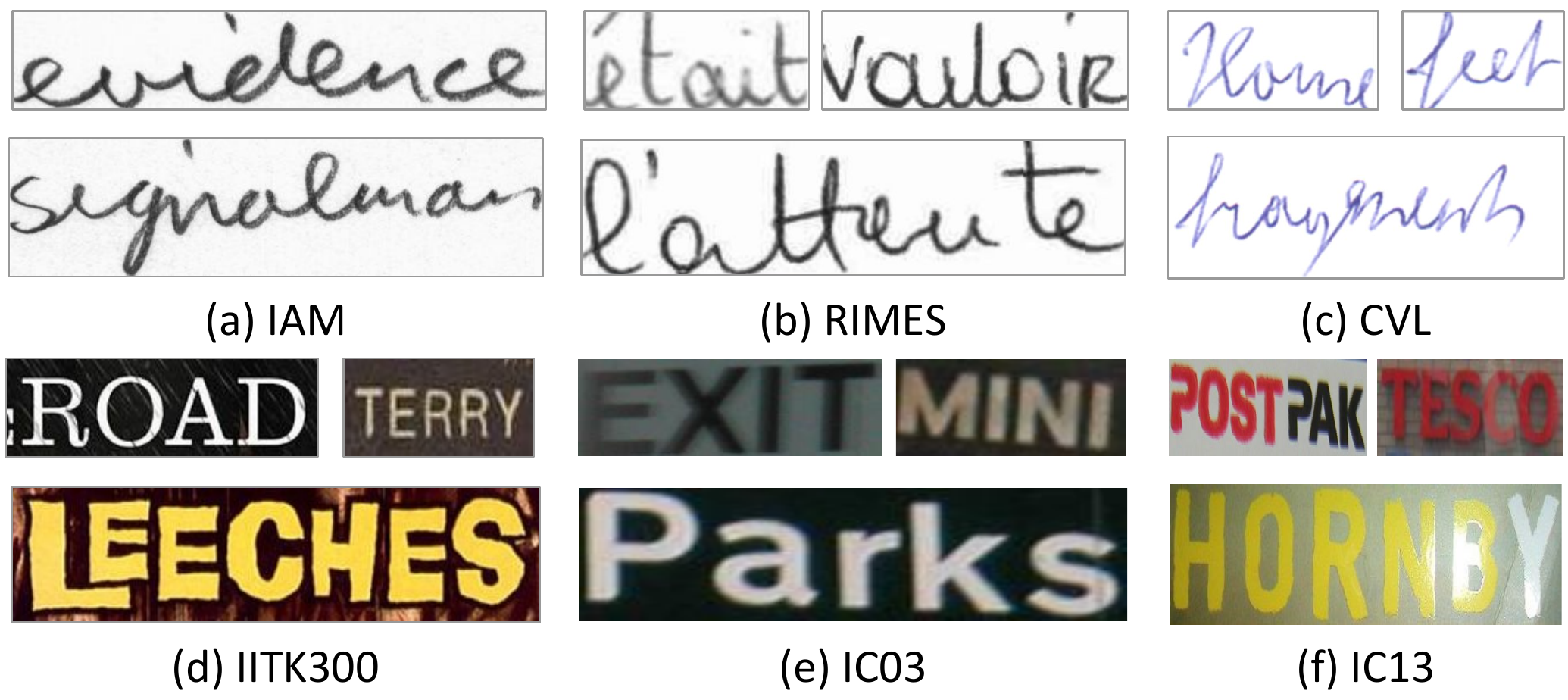}
  \caption{\textbf{Dataset samples.}}
  \label{fig:dataset_samples}
\end{figure}

In this work, we consider the following public datasets for handwriting and scene text, see examples in \cref{fig:dataset_samples}:
\begin{itemize}%[nolistsep]
    \item \textbf{RIMES} \cite{grosicki2009icdar} handwritten French text dataset, written by 1300 different writers, partitioned into writer independent training, validation and test. This collection contains 66,480 correctly segmented words.
    \item \textbf{IAM} \cite{marti2002iam} handwritten English text dataset, written by 657 different writers, partitioned into writer independent training, validation and test. This collection contains 74,805 correctly segmented words.
    \item \textbf{CVL} \cite{kleber2013cvl} handwritten English text dataset, written by 310 different writers, partitioned into writer independent training and test. 27 of the writers wrote 7 texts and the other 283 writers wrote 5 texts. 
    \item \textbf{SynthText} (ST)\cite{gupta2016synthetic} contains 8M cropped scene text images which were generated synthetically. This dataset was utilized for training the scene text recognizer.
    \item \textbf{IIIT5K-words} (IIIT5K)~\cite{Mishra2012sj} contains 2000 training and 3000 testing cropped scene text images from the Internet.
    \item \textbf{ICDAR-2003} (IC03)~\cite{Lucas2003ic03} contains 867 cropped scene text images.
    \item \textbf{ICDAR-2013} (IC13)~\cite{Karatzas2013ic13} contains 848 training and 1015 testing cropped scene text.
\end{itemize}
The last three datasets were used just for validation and test sets as described \cref{app:implementation_details}.

\section{Implementation Details}
\label{app:implementation_details}

\paragraph{Recognizer setting}
Unless otherwise specified, the text recognizer architecture consists of: a feature extraction stage of a 29-layer ResNet~\cite{Kaiming2016resnet}, as in~\cite{cheng2017focusing,baek2019wrong}; a sequence modeling stage using a two-layer Bidirectional-LSTM (BiLSTM) with 256 hidden units per layer; and if needed, an attention decoder of an LSTM cell with 256 memory blocks. 
Following common practice in text recognition \cite{baek2019wrong}, we pre-resize all images to $32\times100$ both for training and testing.
% We use a different symbol classes for each data.
For the English datasets (IAM, CVL, IIIT5K, IC03 and IC13), we use 95 symbol classes: 52 case-sensitive letters, 10 digits and 33 for special characters.
For the French dataset (RIMES), we add to the above the French accent symbols.
As for special symbols for CTC decoding, an additional "[blank]" token is added to the label set.
For the attention decoder, two special symbols are added: “[S]”, “[EOW]” which indicate the start of the sequence and the end of the word.

\paragraph{SimCLR re-implementation}
To compare our method to non-sequential contrastive learning methods, we re-implement the SimCLR algorithm, with the same augmentations and projection head as in \cite{chen2020simple}. This algorithm can be applied in our settings as we anyhow resize each input image to a fixed width following the common practice in text recognition \cite{baek2019wrong,litman2020scatter}.

\paragraph{Pre-training procedure}
In general, for the self-supervised training, we use a batch size of 1024, and train for 200K iterations for handwritten datasets and 400K iterations for scene-text. That said, since frame-to-one mapping results in many more instances for the contrastive loss (Figure 5(c)), we needed to reduce the batch size to 256. To compensate for it, we increased the number of iterations to 300K for handwritten datasets and to 600K for scene-text.
For optimization, we use the AdaDelta optimizer~\cite{zeiler2012adadelta} with a decay rate of 0.95, a gradient clipping parameter with a magnitude of 5 and a weight decay parameter of $10^{-4}$.
The learning rate is initialized to 10, and is reduced by a factor of 10 after 60\% and 80\% of the training iterations.
Finally, all experiments are trained and tested using the PyTorch framework on 4 cards of Tesla V100 GPU with 16GB memory.

\paragraph{Decoder-evaluation and fine-tuning procedures}
For these stages, we train the decoder using a batch size of 256 for 50K iterations, employing a similar learning rate scheduling as in the self-supervised phase. The augmentation procedure consists of light cropping, linear contrast and Gaussian blur. We select our best model using a validation dataset, where in handwritten text we use the public validation sets, and in scene text our validation data is the union of the training data of IC13 and IIIT, as done in \cite{baek2019wrong}.

% \section{Error Vs. Labeled Data Amount}
% \label{app:experiments}

% In \cref{fig:loglog}, we present the word error rate of \AlgoName{} and the \textit{supervised baseline} as a function of the portion of labeled data used for the fine-tuning phase (see Section 5.2). For example, in the RIMES dataset when using an attention decoder, the \AlgoName{} algorithm utilizing $50\%$ of the labeled data achieves the same error rate as training the \textit{supervised baseline} on the entire labeled dataset. 
% In general, as can be seen, \AlgoName{} is effective as almost doubling the size of the labeled dataset.

% {\small
% \bibliographystyle{ieee_fullname}
% \bibliography{bib}
% }

% \end{document}

\end{document}